\documentclass[10pt,twocolumn,letterpaper]{article}

\usepackage{cvpr}              

\usepackage{xspace}
\newcommand{\method}{Cosmos4D\xspace}

\usepackage{multirow}
\usepackage{colortbl}
\definecolor{tabfirst}{RGB}{238,200,112}
\definecolor{tabsecond}{RGB}{250,229,180}
\definecolor{tabthird}{RGB}{232,232,232}
\newcommand{\best}[1]{\cellcolor{tabfirst}\textbf{#1}}
\newcommand{\second}[1]{\cellcolor{tabsecond}#1}
\newcommand{\third}[1]{\cellcolor{tabthird}#1}
\newcommand{\shadebox}[2]{{\setlength{\fboxsep}{1.5pt}\colorbox{#1}{\strut #2}}}
\definecolor{textgold}{RGB}{184,134,11}

\definecolor{cvprblue}{rgb}{0.21,0.49,0.74}
\usepackage[pagebackref,breaklinks,colorlinks,allcolors=cvprblue]{hyperref}

\def\confName{CVPR}
\def\confYear{2027}

\title{See the Lives of Stars in a Handful of Light}

\author{
Shuhong Liu$^{1,2}$ \quad Gengjia Chang$^{2}$ \quad Quanfeng Xu$^{3}$ \quad Ziteng Cui$^{1}$ \quad Qianyu Zhou$^{1}$\\
Xiaoyu Dong$^{1}$ \quad Lin Gu$^{4}$ \quad Tatsuya Harada$^{1,5}$\\[0.6em]
{\normalsize $^{1}$The University of Tokyo \quad $^{2}$I2WM \quad $^{3}$Shanghai Astronomical Observatory}\\
{\normalsize $^{4}$Tohoku University \quad $^{5}$RIKEN AIP}
}

\begin{document}
\pagestyle{plain}
\twocolumn[{%
\renewcommand\twocolumn[1][]{#1}%
\maketitle
\thispagestyle{plain}
\begin{center}
    \centering
    \includegraphics[width=\textwidth]{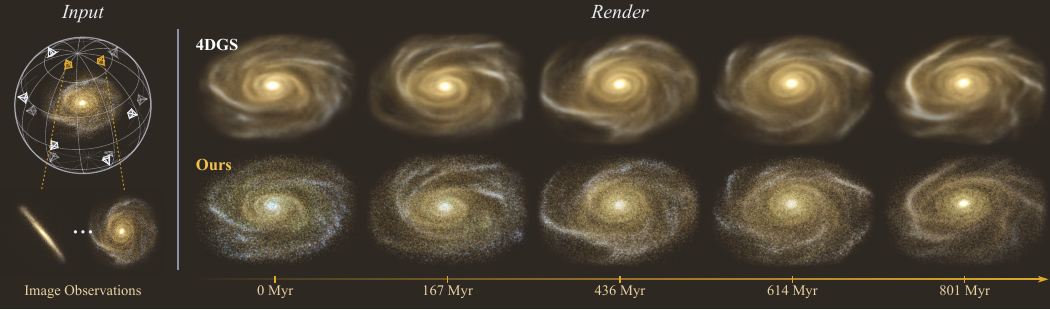}
    \captionof{figure}{\method recovers a galaxy in four dimensions from a handful of multi-view images. Across eight hundred million years, 4DGS~\cite{Wu_2024_CVPR} (top) blurs the stars into a smooth glow, whereas \method (bottom) preserves the star-level texture of the spiral arms.}
    \label{fig:teaser}
\end{center}
\vspace{1em}
}]
\begin{abstract}
Galaxies evolve over hundreds of millions of years, and no telescope has ever watched one move. Cosmological simulations make this evolution visible as a sequence of multi-view images, inviting the question of whether a galaxy can be recovered in four dimensions from its images. This setting defeats existing dynamic scene methods, since millions of stars span only a few pixels each, faint populations blend into diffuse light, and free deformations reproduce the images without obeying gravity. We present Cosmos4D, which rests on two components. Gravitational 4D Gaussian Splatting represents each star as a self-luminous Gaussian whose motion is integrated through a gravitational potential and a low-dimensional force field shared by all stars. Luminosity-stratified dynamics give bright stars individual orbits supervised across image scales and carry faint populations with a velocity field estimated from optical flow. We further build a 4D dataset of eight galaxies from four cosmological simulations, on which Cosmos4D outperforms dynamic Gaussian baselines in both photometric quality and physical accuracy. With these advancements, we glimpse eight hundred million years of a galaxy’s life in ninety-six frames.
\end{abstract}
    
\vspace{-2.5em}
\section{Introduction}
\label{sec:intro}

\begin{flushright}
\begin{minipage}{0.84\linewidth}
\small\itshape
To see a World in a Grain of Sand,\\
Hold Infinity in the palm of your hand.\par
\vspace{3pt}
\upshape\hfill--- William Blake, \emph{Auguries of Innocence}
\end{minipage}
\end{flushright}
\vspace{4pt}

A human life lasts less than a century, while the Sun, moving at about 250\,km\,s$^{-1}$, needs roughly 200 million years to orbit the Milky Way once~\cite{blandhawthorn2016galaxy}. A disk star in Andromeda, 2.5 million light-years away~\cite{li2021m31}, moves at a similar speed~\cite{chemin2009m31}, yet its image shifts by only a seventh of a Hubble pixel per century, which only years of space astrometry or radio interferometry can measure~\cite{brunthaler2005m33,sohn2012m31}. We know that stars sweep along their orbits~\cite{binney2008galactic}, that new stars ignite and older populations redden and fade, yet every image of a galaxy is a single snapshot of this evolution.

Simulations can supply the frames that observations lack. Cosmological simulations such as TNG50~\cite{pillepich2019tng50,nelson2019tng50} follow the stars of a galaxy through cosmic time, allowing it to be rendered from any viewpoint at any moment of its history. The evolution of a galaxy thus becomes a sequence of multi-view images, and a natural question arises. Can a galaxy be recovered in four dimensions from its images alone? The question also poses a demanding test for dynamic reconstruction, because the scene obeys known physical laws while its motion must still be inferred from light.

Such a scene differs in several ways from those that dynamic radiance fields~\cite{Pumarola_2021_CVPR,Wu_2024_CVPR,luiten2024dynamic} were designed for. A galaxy is optically thin and self-luminous, so its hundred thousand point sources add together instead of occluding one another, leaving no surfaces, silhouettes, or depth order to anchor correspondence. Its appearance also changes for reasons unrelated to motion, as stars are born during the sequence and stellar populations brighten, redden, and fade with age. Most importantly, the sequence is long compared with the dynamics. Ninety-six frames span $0.8$\,Gyr, several rotations of the inner disk, whereas the three-dimensional positions of stars are known at only a single instant. A free-form deformation field~\cite{Pumarola_2021_CVPR,park2021nerfies,Wu_2024_CVPR} can then fit every frame with motion that no galaxy could follow. Per-star trajectories~\cite{luiten2024dynamic} drift away, since a star spanning only a few pixels receives no image gradient once it moves off its target.

The problem becomes tractable once we recall that stars do not move independently. They fall together through a shared gravitational potential~\cite{miyamoto1975disk,hernquist1990bulge,navarro1997nfw}, which is shaped by the very distribution of mass observed at the known instant~\cite{binney2008galactic}. Their motion is therefore governed by an equation of motion rather than by an arbitrary function of space and time, and the trajectories it permits form a far smaller family than the deformations a network can express. Gravity should nevertheless guide the reconstruction without demanding more precision than images can offer. Over hundreds of millions of years the identity of an individual star cannot be recovered from images, and faint populations leave almost no trace of their own motion.

Guided by these observations, we present \method, which represents an evolving galaxy as a set of self-luminous Gaussians~\cite{kerbl2023gaussians} driven by gravity. Each star becomes a Gaussian rendered by additive splatting, with its changing emission separated from its motion. Its position follows a differentiable orbit integrator~\cite{chen2018neuralode} through a potential inferred from the known instant, corrected by a low-dimensional force field shared by all stars. Because images constrain stars unequally, bright stars receive individual orbital states, while faint stars, which stand for unresolved populations, are carried by a shared velocity field estimated from multi-view optical flow. Because a bright star renders as a point source only a few pixels wide, we also supervise it across an image pyramid, which keeps guiding it toward its target after it drifts beyond its own footprint. To evaluate this setting, we build a dataset of eight evolving galaxies from four cosmological simulations~\cite{pillepich2019tng50,nelson2019tng50} and measure both the photometric quality of the renderings and the physical accuracy of the recovered motion.

To summarize, our contributions are as follows.
\begin{itemize}
\item We propose \method, a Gravitational 4D Gaussian Splatting framework that models each star as a self-luminous Gaussian whose motion is integrated through a gravitational potential and a low-dimensional force field shared by all stars.
\item We introduce luminosity-stratified dynamics, which give bright stars individual orbits supervised across an image pyramid and carry faint populations with a velocity field estimated from images.
\item We build a multi-view dataset of eight evolving galaxies from four cosmological simulations, on which \method outperforms dynamic Gaussian baselines in both rendering fidelity and physical accuracy.
\end{itemize}

\begin{figure*}[t]
\centering
\includegraphics[width=\textwidth]{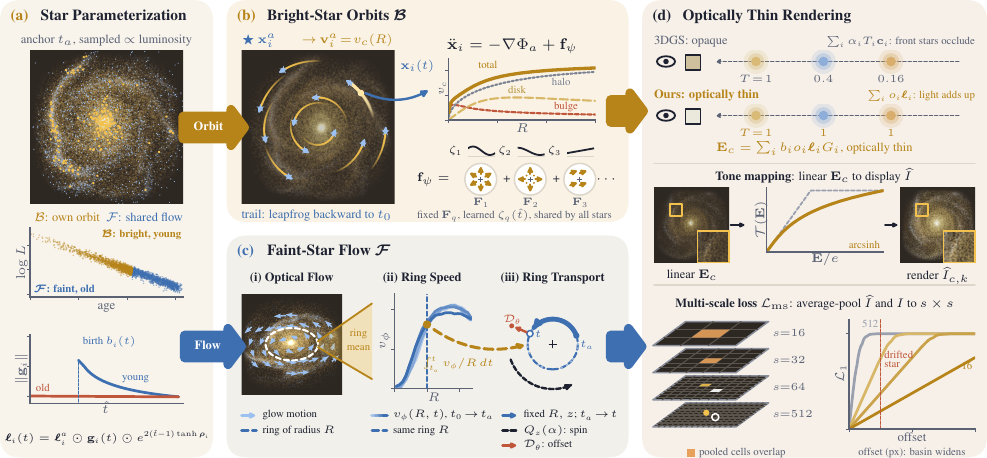}
\caption{Overview of \method. (a) Anchor stars are split by luminosity into young bright stars $\mathcal{B}$ and old faint stars $\mathcal{F}$, each a self-luminous Gaussian with evolving emission. (b) Each bright star follows its own orbit, integrated backward under the fixed potential $\Phi_a$ and a shared learned force $\mathbf{f}_\psi$. (c) Faint stars circle their rings at the speed measured from optical flow, refined by a shared spin $Q_z(\alpha)$ and an offset $\mathcal{D}_\theta$. (d) Stars add their light rather than occlude, and average pooling lets even drifted stars still overlap their observed targets.}
\label{fig:pipeline}
\end{figure*}

\section{Related Work}
\label{sec:related_work}

\subsection{Dynamic Scene Representations}
\label{sec:rw_dynamic}

Building on NeRF~\cite{mildenhall2020nerf}, canonical deformation~\cite{Pumarola_2021_CVPR,park2021nerfies} and higher-dimensional embeddings~\cite{park2021hypernerf} accommodate nonrigid motion and topology changes. Factorized fields~\cite{Cao_2023_CVPR,fridovich2023kplanes} improve spatiotemporal efficiency, while 3D Gaussian Splatting~\cite{kerbl2023gaussians} enables fast explicit rendering. Its dynamic extensions encode temporal correspondence through persistent particles~\cite{luiten2024dynamic}, shared deformation fields~\cite{Wu_2024_CVPR,Yang_2024_CVPR}, or parametric spacetime primitives~\cite{li2024spacetime}, trading motion flexibility against compactness. This progression replaces independent frame reconstruction with temporal parameter sharing, enabling coherent interpolation while limiting the cost of representing long sequences.

Sparse controls and motion scaffolds~\cite{huang2024scgs,Lei_2025_CVPR} constrain neighboring trajectories, and recent motion bases~\cite{Wang_2025_ICCV,Zhou_2026_CVPR,zhang2026continuous} further structure reconstruction across time and space. These advances improve temporal coherence, but their geometric and kinematic priors do not specify stellar accelerations or population luminosity evolution. In optically thin stellar systems, changing emission and unequal motion evidence across populations require additional structure. \method couples evolving emission with gravitational orbits for bright stars and a shared flow for faint stars.

\subsection{Physics-Informed Reconstruction}
\label{sec:rw_physics}

Neural ODEs~\cite{chen2018neuralode} learn continuous dynamics, while Hamiltonian models~\cite{greydanus2019hamiltonian}, equation residuals~\cite{raissi2019pinn}, and differentiable simulators~\cite{hu2020difftaichi} incorporate physical structure. Coupled with rendering, these ideas support fluid reconstruction~\cite{chu2022physics,du2025gaussfluids}, deformable system identification~\cite{qiao2022neuphysics,li2023pacnerf}, and spring-mass estimation~\cite{zhong2024springgaus,jiang2025phystwin}. Shared simulation--rendering primitives~\cite{xie2024physgaussian} additionally connect reconstructed Gaussian assets to material-based animation. In inverse formulations, physical structure constrains hidden quantities that images alone leave ambiguous, including geometry and motion.

Complementary approaches infer simulation parameters from video-generation priors~\cite{zhang2024physdreamer,liu2025physflow} or multimodal material cues~\cite{zhao2025physsplat}, while recent physical Gaussian prediction~\cite{lv2026physgm} and learned force fields~\cite{li2026ngff} extend this direction toward efficient dynamics synthesis. These methods primarily address material response and object interactions. Our task instead reconstructs stellar motion from images and a single positional anchor, requiring gravitational acceleration coupled to evolving emission. We allocate orbital states according to population luminosity and supervise point-like stars across image scales so that long trajectories remain optimizable.

\section{Method}
\label{sec:method}

\method reconstructs an evolving galaxy from multi-view images, representing each star as a self-luminous Gaussian that moves along a gravitational orbit. We first introduce Gravitational 4D Gaussian Splatting (\cref{sec:gravity_gaussians}), then Luminosity-Stratified Dynamics (\cref{sec:hybrid_dynamics}), which assigns orbital freedom according to stellar brightness, and finally the training objective (\cref{sec:optimization}). \cref{fig:pipeline} gives an overview.

\paragraph{Problem Formulation.}
Given calibrated images $\{I_{c,k}\}$ from cameras $c$ at snapshot times $\{t_k\}_{k=0}^{T-1}$, we aim to reconstruct the evolving appearance of a galaxy together with physically plausible orbits for its stars. Since a real galaxy can be seen at just one moment of its history, we assume that its stars are known only at the latest snapshot $t_a=t_{T-1}$. This snapshot, which we call the \textbf{\emph{anchor}}, plays the role of the present day. It provides the position $\mathbf{x}_i^a$, age, metallicity, and luminosity of each star $i$, as well as the total stellar mass $M_\star$. Velocities, individual masses, and all earlier configurations remain unknown. The whole scene evolves in cosmic time $t$ (Gyr). We also write $\hat t=k/(T-1)$ for the normalized time, which runs from 0 at the first snapshot to 1 at the anchor.

\subsection{Gravitational 4D Gaussian Splatting}
\label{sec:gravity_gaussians}

\paragraph{Optically Thin Rendering.}
We represent each sampled star by one isotropic Gaussian with a time-dependent center $\mathbf{x}_i(t)$, a learned radius, an opacity $o_i$, and a view-independent RGB emission $\boldsymbol{\ell}_i(t)$. Isotropy keeps each star a point source and prevents elongated splats from absorbing motion errors. The number of Gaussians is fixed, with no densification or pruning.

We treat the galaxy as optically thin and free of dust, so stars add their light rather than occlude one another. The linear RGB image rendered for camera $c$ is thus a plain sum,
\begin{equation}
\mathbf{E}_c(\mathbf{u},t)=
\sum_i b_i(t)o_i\boldsymbol{\ell}_i(t)
\exp\!\left[-\tfrac12\mathbf{d}_{ic}^{\top}
\boldsymbol{\Sigma}_{ic}^{-1}\mathbf{d}_{ic}\right],
\label{eq:emission}
\end{equation}
where $\mathbf{u}$ is a pixel location, $\mathbf{d}_{ic}=\mathbf{u}-\boldsymbol{\mu}_{ic}(t)$, $\boldsymbol{\mu}_{ic}$ and $\boldsymbol{\Sigma}_{ic}$ are the projected mean and covariance, and the birth mask $b_i(t)=\mathbf{1}[t\geq t_i^{\mathrm{birth}}]$ hides each star until it forms. A custom differentiable rasterizer evaluates this sum and passes image gradients to both appearance and positions.

Stellar surface brightness drops by several orders of magnitude from the galactic center to the outer disk~\cite{freeman1970disks}. We therefore pass the linear image $\mathbf{E}_c$ of \cref{eq:emission} through the arcsinh tone mapping $\mathcal{T}$ commonly used in astronomical imaging~\cite{lupton2004rgb}, which stays linear for faint light and compresses the bright cores. This gives the rendered image $\widehat I_{c,k}=\mathcal{T}(\mathbf{E}_c(\cdot,t_k))$, which is optimized to match the observation $I_{c,k}$. Appendix~\ref{sec:suppl_appearance} details the tone mapping.

\paragraph{Evolving Emission.}
Stars change in brightness and color as they age. Young stars are blue and fade within tens of millions of years, while old stars are red and change slowly. To separate this evolution from stellar motion, we write the emission of each star as a product of three terms,
\begin{equation}
\boldsymbol{\ell}_i(t)=\boldsymbol{\ell}_i^a\odot
\mathbf{g}_i(t)\odot
\exp\!\left[2(\hat t-1)\tanh\boldsymbol{\rho}_i(\hat t)\right],
\label{eq:luminosity}
\end{equation}
where $\odot$ is element-wise multiplication over the RGB channels and $\hat t$ is the normalized frame time. The first term $\boldsymbol{\ell}_i^a$ is the learned RGB emission of star $i$ at the anchor, initialized from its age, metallicity, and luminosity. The second term $\mathbf{g}_i(t)$ is a fixed per-channel gain that scales the anchor emission back in time. We compute $\mathbf{g}_i(t)$ from a simple analytic stellar population model, in which a star fades as a power law of its age and reddens as it grows older. Young stars thus appear brighter and bluer in early frames. The third term is a learned correction for what the analytic model misses, where $\boldsymbol{\rho}_i(\hat t)$ is a per-star RGB curve over the normalized time. The $\tanh$ bounds this correction, so image changes caused by stellar motion are left to the dynamics.

\paragraph{Gravitational Dynamics.}
Stars in a galaxy move under the combined gravity of stars, gas, and dark matter~\cite{binney2008galactic}. Instead of learning a free displacement per star and frame, we let each star follow an orbit whose acceleration depends on where the star is,
\begin{equation}
\begin{aligned}
\ddot{\mathbf{x}}_i(t)&=\mathbf{a}_\psi(\mathbf{x}_i(t),t),\\
\mathbf{a}_\psi(\mathbf{x},t)&=-\nabla\Phi_a(\mathbf{x})+
\mathbf{f}_\psi(\mathbf{x},t),
\end{aligned}
\label{eq:dynamics}
\end{equation}
where $\ddot{\mathbf{x}}_i$ is the acceleration of star $i$, and positions are expressed in a frame centered on the galaxy and aligned with its disk. The acceleration combines the pull of a fixed gravitational potential $\Phi_a$ with a learned residual force $\mathbf{f}_\psi$. The potential acts as an energy landscape whose downhill direction $-\nabla\Phi_a$ points toward the attracting mass. We build $\Phi_a$ from the anchor stars and $M_\star$ as a standard analytic model of a central bulge, a flat disk, and a dark matter halo~\cite{hernquist1990bulge,miyamoto1975disk,navarro1997nfw}. This smooth model is axisymmetric, so it cannot capture bars, spiral arms, or deviations from the assumed mass distribution~\cite{sellwood2014secular}. We therefore learn the residual force $\mathbf{f}_\psi$ to account for such unmodeled forces,
\begin{equation}
\mathbf{f}_\psi(\mathbf{x},t)=
\sum_{q} \zeta_q(\hat t)\,\mathbf{F}_q(\mathbf{x}),
\label{eq:force_basis}
\end{equation}
where each $\mathbf{F}_q$ is a predefined smooth force field and $\zeta_q(\hat t)$ is the learned time-varying coefficient of that field. All stars share these coefficients, so they follow one common force rather than per-star offsets.

To obtain the trajectory of a star, we start from its orbital state at the anchor, namely the position $\mathbf{x}_i^a$ and velocity $\mathbf{v}_i^a$, and integrate the acceleration of \cref{eq:dynamics} twice backward in time with a differentiable leapfrog scheme~\cite{binney2008galactic}. The anchor position is known, and the velocity is learned from the images. Image losses pass gradients through every integration step to the orbital states and the force coefficients. Appendix~\ref{sec:suppl_dynamics} details the potential, force fields, and integrator.

\begin{figure}[t]
\centering
\includegraphics[width=\linewidth]{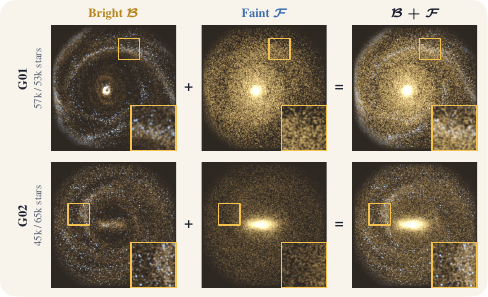}
\caption{Luminosity stratification. Bright stars $\mathcal{B}$ are distinct points along the spiral arms, while faint stars $\mathcal{F}$ blend into a glow.}
\label{fig:strat}
\end{figure}

\begin{figure*}[t]
    \centering
    \includegraphics[width=\textwidth]{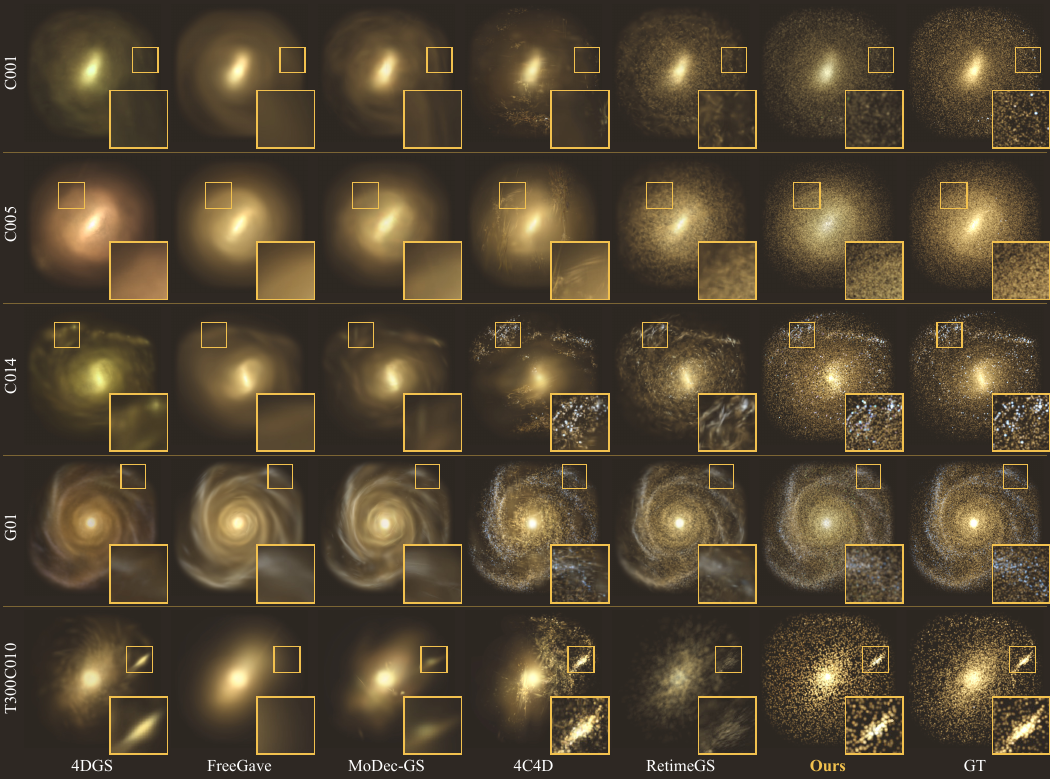}
    \caption{Qualitative comparison on test views, with the boxed region magnified at the bottom-right of each tile. Baselines blur the stellar disk or break it into streaks, whereas \method recovers star-level texture, young blue stars, and tidal features seen in GT.}
    \label{fig:qualitative}
\end{figure*}
\subsection{Luminosity-Stratified Dynamics}
\label{sec:hybrid_dynamics}

Representing every star of a galaxy as a Gaussian is computationally prohibitive, so we model the galaxy with an importance sample of its stars, drawn with probability proportional to luminosity. At the anchor, we stratify the sampled stars by luminosity into a bright set $\mathcal{B}$ and a faint set $\mathcal{F}$. Every bright star is kept and modeled individually, whereas each faint star stands in for several similar neighbors. As shown in \cref{fig:strat}, each bright star appears as a distinct spot, while faint stars blend into a smooth glow. We therefore assign each bright star its own gravitational orbit, while faint stars share one flow field estimated from the images. Both groups keep the evolving emission of \cref{eq:luminosity}.

\paragraph{Bright-Star Orbits.}
Stars in $\mathcal{B}$ follow the gravitational dynamics of \cref{eq:dynamics}. Their velocities are initialized to the circular speed $v_c(R)=\sqrt{R\,\partial\Phi_a/\partial R}$ of the anchor potential at their anchor radius $R$, and are then refined during training. Each bright star is labeled as a disk or halo star from its height above the disk plane. Disk stars start on circular orbits around the disk axis, matching the rotation of a disk galaxy, while halo stars start on circular orbits in randomly oriented planes.

\paragraph{Faint-Star Flow.}
Although individual faint stars cannot be tracked, their combined glow shows how the disk rotates. We therefore fit a shared velocity field for $\mathcal{F}$ to the optical flow of the training views. Image patterns such as spiral arms can travel at a speed different from that of their stars, so we keep only the rotation speed $v_\phi(R,t)$ at each radius $R$. Each faint star circles the disk axis at its anchor radius $R_i^a$ and height $z_i^a$. In disk-centered cylindrical coordinates $(R,\phi,z)$, we integrate
\begin{equation}
\dot\phi_i(t)=\frac{v_\phi(R_i^a,t)}{R_i^a},\qquad
R_i(t)=R_i^a,\quad z_i(t)=z_i^a,
\label{eq:transport}
\end{equation}
backward in time from the anchor, which gives the flow position $\mathbf{x}_i^{\mathrm{flow}}(t)$. The radial and vertical profiles of the faint population thus stay exactly as observed at the anchor. We refine this pure rotation with two bounded corrections,
\begin{equation}
\mathbf{x}_i(t)=Q_z(\alpha(\hat t))\,\mathbf{x}_i^{\mathrm{flow}}(t)
+(\hat t-1)\,\mathcal{D}_\theta(\mathbf{x}_i^a,\hat t),
\label{eq:faint_residual}
\end{equation}
where $Q_z(\alpha)$ rotates the whole faint population about the disk axis by a shared angle $\alpha(\hat t)$, and each star is further moved by a small offset. This offset comes from a deformation network $\mathcal{D}_\theta$ with parameters $\theta$, which follows 4DGS~\cite{Wu_2024_CVPR} and combines a HexPlane encoding~\cite{Cao_2023_CVPR} with a small MLP. The network takes the anchor position of a star and the normalized time as input and predicts a bounded 3D offset, so a faint star moves only a limited distance away from its flow position. The factor $(\hat t-1)$ makes the offset vanish at the anchor, where the stars are observed. Details of the flow estimation are given in Appendix~\ref{sec:suppl_flow}.

\begin{table*}[t]
  \centering
  \caption{Quantitative comparison of novel-view rendering on all eight scenes. For each scene, Mask reports metrics within the galaxy foreground mask and Full reports metrics on the whole image. All metrics are averaged over three test views and 96 frames at $512^2$ resolution. Darker cells mark the best, second best, and third best results.}
  \label{tab:main}
  \footnotesize
  \def\tabmainbody{%
    \toprule
    \multirow{2}{*}{\textbf{Methods}} & \multirow{2}{*}{\textbf{Metrics}} & \multicolumn{2}{c}{G01} & \multicolumn{2}{c}{G02} & \multicolumn{2}{c}{C001} & \multicolumn{2}{c}{C010} & \multicolumn{2}{c}{C005} & \multicolumn{2}{c}{C014} & \multicolumn{2}{c}{T300C010} & \multicolumn{2}{c|}{T300C021} & \multicolumn{2}{c}{\textbf{Avg.}} \\
    \cmidrule(lr){3-4} \cmidrule(lr){5-6} \cmidrule(lr){7-8} \cmidrule(lr){9-10} \cmidrule(lr){11-12} \cmidrule(lr){13-14} \cmidrule(lr){15-16} \cmidrule(lr){17-18} \cmidrule(lr){19-20}
    & & Mask & Full & Mask & Full & Mask & Full & Mask & Full & Mask & Full & Mask & Full & Mask & Full & Mask & Full & Mask & Full \\
    \midrule
    \multirow{3}{*}{4DGS~\cite{Wu_2024_CVPR}} & PSNR$\uparrow$ & 19.10 & 25.15 & 20.27 & 25.42 & \second{20.41} & \second{26.00} & \best{24.42} & \best{29.45} & 23.59 & \third{29.77} & 19.91 & 25.47 & \third{17.15} & \second{24.48} & 15.03 & 23.10 & 19.99 & 26.11 \\
     & SSIM$\uparrow$ & 0.302 & 0.670 & 0.348 & 0.632 & 0.286 & 0.603 & 0.391 & 0.610 & 0.458 & 0.654 & 0.322 & 0.627 & 0.149 & 0.554 & 0.110 & \third{0.643} & 0.296 & 0.624 \\
     & LPIPS$\downarrow$ & 0.896 & 0.576 & 0.857 & 0.561 & 0.870 & 0.594 & 0.763 & 0.590 & 0.671 & 0.489 & 0.812 & 0.575 & 0.792 & 0.679 & 0.805 & 0.690 & 0.808 & 0.594 \\
    \midrule
    \multirow{3}{*}{FreeGave~\cite{li2025freegave}} & PSNR$\uparrow$ & 19.99 & 26.00 & 20.75 & 25.88 & 19.77 & 25.21 & \second{24.29} & \second{29.40} & \third{23.91} & \second{30.14} & 19.86 & 25.45 & 17.08 & 24.38 & \second{15.62} & \second{24.80} & \third{20.16} & \second{26.41} \\
     & SSIM$\uparrow$ & 0.320 & 0.685 & 0.362 & 0.660 & 0.283 & \third{0.622} & \third{0.405} & \third{0.643} & \second{0.493} & \best{0.693} & 0.349 & 0.659 & 0.168 & 0.556 & \third{0.115} & \best{0.645} & 0.312 & 0.645 \\
     & LPIPS$\downarrow$ & 0.926 & 0.586 & 0.866 & 0.567 & 0.886 & 0.600 & 0.755 & 0.586 & 0.659 & 0.484 & 0.820 & 0.578 & 0.818 & 0.686 & 0.820 & 0.695 & 0.819 & 0.598 \\
    \midrule
    \multirow{3}{*}{MoDec-GS~\cite{kwak2025modecgs}} & PSNR$\uparrow$ & \third{20.01} & \third{26.03} & \third{21.10} & \third{26.24} & 19.83 & 25.41 & 22.86 & 28.13 & \second{23.92} & \best{30.20} & \second{20.26} & \second{25.85} & \second{17.16} & \best{24.49} & 15.23 & \third{24.43} & 20.05 & \third{26.35} \\
     & SSIM$\uparrow$ & 0.324 & 0.686 & 0.380 & 0.668 & 0.282 & 0.610 & 0.400 & 0.632 & \third{0.481} & 0.675 & 0.366 & \second{0.666} & \third{0.169} & \second{0.560} & 0.104 & \third{0.643} & 0.313 & 0.643 \\
     & LPIPS$\downarrow$ & 0.887 & 0.572 & 0.844 & 0.557 & 0.868 & 0.593 & 0.761 & 0.588 & 0.660 & 0.485 & 0.797 & 0.570 & 0.804 & 0.683 & 0.807 & 0.687 & 0.804 & 0.592 \\
    \midrule
    \multirow{3}{*}{4C4D~\cite{zhou2026fourc4d}} & PSNR$\uparrow$ & 19.70 & 25.74 & \second{21.87} & \second{27.26} & 20.24 & 25.75 & 23.46 & 28.82 & 23.58 & 28.75 & \best{20.29} & \second{25.85} & \best{17.17} & \third{24.46} & \best{15.78} & \best{24.96} & \second{20.26} & \best{26.45} \\
     & SSIM$\uparrow$ & \best{0.391} & \third{0.705} & \third{0.402} & \third{0.679} & \third{0.308} & 0.611 & \second{0.408} & 0.633 & 0.480 & \third{0.678} & \third{0.386} & \third{0.662} & \second{0.175} & \third{0.557} & \second{0.120} & \second{0.644} & \second{0.334} & \third{0.646} \\
     & LPIPS$\downarrow$ & \second{0.486} & \second{0.445} & \best{0.602} & \second{0.474} & \second{0.652} & \second{0.509} & \third{0.696} & \third{0.563} & \third{0.641} & \third{0.477} & \second{0.593} & \second{0.488} & \second{0.677} & \second{0.648} & \third{0.769} & \third{0.682} & \second{0.640} & \third{0.536} \\
    \midrule
    \multirow{3}{*}{RetimeGS~\cite{wang2026retimegs}} & PSNR$\uparrow$ & \second{20.12} & \second{26.14} & 20.89 & 26.02 & \third{20.27} & \third{25.86} & \third{23.82} & \third{29.09} & \best{24.04} & 29.34 & \best{20.29} & \best{25.89} & 16.50 & 23.91 & 14.62 & 23.85 & 20.07 & 26.26 \\
     & SSIM$\uparrow$ & \third{0.382} & \best{0.714} & \second{0.405} & \best{0.694} & \second{0.343} & \second{0.638} & \best{0.429} & \best{0.647} & \best{0.498} & \second{0.691} & \second{0.389} & 0.661 & 0.130 & 0.549 & 0.091 & 0.638 & \third{0.333} & \best{0.654} \\
     & LPIPS$\downarrow$ & \third{0.656} & \third{0.482} & \second{0.618} & \best{0.455} & \third{0.669} & \third{0.515} & \second{0.597} & \second{0.527} & \second{0.538} & \second{0.442} & \third{0.613} & \third{0.500} & \third{0.729} & \third{0.666} & \second{0.740} & \second{0.681} & \third{0.645} & \second{0.534} \\
    \midrule
    \multirow{3}{*}{\textbf{Ours}} & PSNR$\uparrow$ & \best{20.44} & \best{26.45} & \best{22.40} & \best{27.45} & \best{20.84} & \best{26.38} & 23.29 & 28.55 & 23.53 & 28.83 & \third{20.20} & \third{25.52} & 17.01 & 23.74 & \third{15.38} & 23.54 & \best{20.39} & 26.31 \\
     & SSIM$\uparrow$ & \second{0.387} & \second{0.712} & \best{0.409} & \second{0.683} & \best{0.377} & \best{0.663} & 0.404 & \second{0.644} & 0.416 & 0.655 & \best{0.400} & \best{0.673} & \best{0.183} & \best{0.561} & \best{0.142} & 0.629 & \best{0.340} & \second{0.653} \\
     & LPIPS$\downarrow$ & \best{0.363} & \best{0.330} & \third{0.705} & \third{0.487} & \best{0.476} & \best{0.377} & \best{0.439} & \best{0.418} & \best{0.230} & \best{0.262} & \best{0.279} & \best{0.276} & \best{0.330} & \best{0.387} & \best{0.316} & \best{0.381} & \best{0.392} & \best{0.365} \\
    \bottomrule}
  \setlength{\tabcolsep}{0pt}
  \settowidth{\dimen0}{\begin{tabular}{l|l|cccccccccccccccc|cc}\tabmainbody\end{tabular}}
  \setlength{\tabcolsep}{\dimexpr(\textwidth-\dimen0)/40\relax}
  \begin{tabular}{l|l|cccccccccccccccc|cc}\tabmainbody\end{tabular}
\end{table*}
\subsection{Optimization}
\label{sec:optimization}

\paragraph{Multi-Scale Supervision.}
Each bright star projects to a compact footprint of only a few pixels. When its position error exceeds this extent, the rendered star no longer overlaps its observed counterpart, and the per-pixel loss provides no gradient toward the correct position. We therefore compare the rendered image $\widehat I_{c,k}$ and the observation $I_{c,k}$ over an image pyramid,
\begin{equation}
\begin{aligned}
\mathcal{L}_{\mathrm{ms}}&=\sum_{s\in\mathcal{S}}
\mathcal{L}_{1}\big(\mathcal{P}_s(\widehat I_{c,k}),\,\mathcal{P}_s(I_{c,k})\big),\\
\mathcal{L}_{1}(\widehat I,I)&=\frac{\sum_{\mathbf{u}}w(\mathbf{u})\,\|\widehat I(\mathbf{u})-I(\mathbf{u})\|_1}{\sum_{\mathbf{u}}w(\mathbf{u})},
\end{aligned}
\label{eq:pyramid}
\end{equation}
where $\mathcal{P}_s$ average-pools an image to $s\times s$ pixels, and $\mathcal{S}=\{512,64,32,16\}$ covers the full image resolution and three coarse levels. The distance $\mathcal{L}_{1}$ averages the RGB error over pixels $\mathbf{u}$ with a pixel weight $w(\mathbf{u})$ that grows with the observed brightness $I(\mathbf{u})$, so the galaxy is not outweighed by the dark sky around it. Pooling enlarges the footprint of each star relative to the pixel grid, which widens the basin of convergence so that misaligned stars still receive informative gradients. The full-resolution level then refines the positions of these stars.

\paragraph{Overall Loss.}
We jointly optimize the appearance of all stars, the orbital states of bright stars, the shared force coefficients, and the faint-star corrections by minimizing
\begin{equation}
\mathcal{L}=\mathcal{L}_{\mathrm{ms}}+\lambda_{\mathrm{ssim}}\mathcal{L}_{\mathrm{ssim}}+\mathcal{L}_{\mathrm{reg}},
\label{eq:training_objective}
\end{equation}
where $\mathcal{L}_{\mathrm{ms}}$ is the multi-scale image loss of \cref{eq:pyramid}, and $\lambda_{\mathrm{ssim}}$ weights the structural loss $\mathcal{L}_{\mathrm{ssim}}=1-\mathrm{SSIM}(\widehat I_{c,k},I_{c,k})$~\cite{wang2004ssim}, which compares local patterns such as spiral arms and star-forming clumps. The regularizer
\begin{equation}
\mathcal{L}_{\mathrm{reg}}=\sum_{\xi\in\{\psi,\,\boldsymbol{\rho}\}}\lambda_\xi\left(\|\xi\|_2^2+\gamma_\xi\|\Delta_t\xi\|_2^2\right)
\label{eq:regularizer}
\end{equation}
acts on the force coefficients $\psi=\{\zeta_q\}$ of \cref{eq:force_basis} and the luminosity correction $\boldsymbol{\rho}$ of \cref{eq:luminosity}. The squared norm keeps each correction small, and the squared difference $\Delta_t\xi$ between neighboring time samples keeps it smooth, with weights $\lambda_\xi$ and $\gamma_\xi$. Details are given in Appendix~\ref{sec:suppl_impl}.

\section{Experiments}
\label{sec:experiment}

\paragraph{Implementation Details.}
Training has two stages. An anchor-only stage of 300 iterations fits color, opacity, and splat radii with every star held at its anchor position, after which the radii are frozen. Joint optimization of dynamics and appearance then runs for 32,000 Adam iterations with batch size one, with a 400-iteration linear warmup followed by cosine decay. We set $\lambda_{\mathrm{ssim}}=0.2$. Further settings are given in Appendix~\ref{sec:suppl_impl}.

\paragraph{Dataset.}
We build eight evolving galaxies from the subboxes of the TNG50~\cite{pillepich2019tng50,nelson2019tng50}, TNG100 and TNG300~\cite{pillepich2018tng,springel2018tng,nelson2018tng}, and Illustris~\cite{vogelsberger2014illustris,genel2014illustris} cosmological simulations, using the last 96 snapshots of each sequence. Stellar particles are importance-sampled by luminosity with identities fixed across frames and rendered as optically thin emitters into 14 cameras on a sphere at $512^2$ resolution. Each scene has 11 training and 3 test views, detailed in Appendix~\ref{sec:suppl_data}.

\paragraph{Baselines \& Metrics.}
We compare \method with 4DGS~\cite{Wu_2024_CVPR}, FreeGave~\cite{li2025freegave}, MoDec-GS~\cite{kwak2025modecgs}, 4C4D~\cite{zhou2026fourc4d}, and RetimeGS~\cite{wang2026retimegs} on all eight scenes and evaluate novel-view rendering with PSNR, SSIM~\cite{wang2004ssim}, and LPIPS~\cite{zhang2018lpips}. Each metric is computed on the full image and within the galaxy foreground mask, which keeps pixels whose ground-truth intensity exceeds $0.02$. Because dark sky dominates the frame, the masked scores better reflect the quality of the galaxy itself. PSNR and SSIM are known to favor blurred over sharp predictions~\cite{blau2018perception}, a bias amplified for stars a few pixels wide, since a slightly misplaced star is penalized at both its true and its predicted location. LPIPS compares deep features and better reflects resolved stars. Best results are shaded as \shadebox{tabfirst}{\textbf{first}}, \shadebox{tabsecond}{second}, and \shadebox{tabthird}{third}.

\subsection{Photometric Evaluation}
\label{sec:photometric}
\cref{tab:main} summarizes rendering quality. \method attains the best average LPIPS in both the masked and full-image settings and the best masked PSNR and SSIM, and the margin is widest in LPIPS, where our masked average of $0.392$ is $39\%$ lower than the $0.640$ of 4C4D and at least $51\%$ lower than the deformation-based methods. \method ranks first in LPIPS on seven of eight scenes, except the dense G02, where splats spanning several pixels blur the core, and leads masked PSNR on G01, G02, and C001. On C010, C005, whose faint particles follow the shared flow instead of orbits, C014, and the sparse TNG300 scenes, baselines reach higher PSNR through smooth predictions that per-pixel metrics favor, whereas \method resolves the stars.

\cref{fig:qualitative} illustrates why LPIPS is more informative. 4DGS, FreeGave, and MoDec-GS render each galaxy as a smooth glow without resolved stars, while 4C4D and RetimeGS stretch their Gaussians into streaks along the motion, as in the outskirts of C001 and C005. \method keeps stars compact across the disk, preserves the young blue population of C014, and reproduces the tidal feature in T300C010 that the other methods miss or smear.

\subsection{Physical Analysis}
\label{sec:physical}
\begin{table}[t]
  \centering
  \caption{Relative error of the rotation curve $v_\phi(R)$ $\downarrow$, measured on the 3D Gaussian centers exported from each model.}
  \label{tab:vphi}
  \scriptsize
  \def\tabphysbody{%
    \toprule
    Scene & 4DGS & FreeGave & MoDec-GS & 4C4D & RetimeGS & Ours \\
    \midrule
    G01 & 0.984 & 0.921 & 0.960 & \third{0.550} & \second{0.468} & \best{0.121} \\
    G02 & 0.987 & 0.962 & 0.969 & \third{0.784} & \best{0.512} & \second{0.522} \\
    C001 & 0.990 & 0.958 & 0.969 & \third{0.704} & \second{0.471} & \best{0.016} \\
    C010 & 0.996 & 0.979 & 0.986 & \third{0.965} & \second{0.849} & \best{0.096} \\
    C005 & 0.995 & 0.982 & 0.973 & \third{0.910} & \second{0.689} & \best{0.677} \\
    C014 & 0.984 & 0.854 & 0.933 & \third{0.723} & \second{0.402} & \best{0.267} \\
    T300C010 & 0.975 & \third{0.876} & 0.925 & \best{0.706} & 0.926 & \second{0.776} \\
    T300C021 & 0.940 & 0.919 & 0.959 & \second{0.569} & \third{0.729} & \best{0.383} \\
    \midrule
    Avg. & 0.981 & 0.931 & 0.959 & \third{0.739} & \second{0.631} & \best{0.357} \\
    \bottomrule}
  \setlength{\tabcolsep}{0pt}
  \settowidth{\dimen0}{\begin{tabular}{lcccccc}\tabphysbody\end{tabular}}
  \setlength{\tabcolsep}{\dimexpr(\columnwidth-\dimen0)/14\relax}
  \begin{tabular}{lcccccc}\tabphysbody\end{tabular}
\end{table}

\begin{table}[t]
  \centering
  \caption{Angular momentum conservation error $\downarrow$ in dex. RetimeGS primitives stay active for only about four frames, so their $L_z$ along a trajectory is undefined and is thus reported as a dash.}
  \label{tab:lz}
  \scriptsize
  \def\tabphysbody{%
    \toprule
    Scene & 4DGS & FreeGave & MoDec-GS & 4C4D & RetimeGS & Ours \\
    \midrule
    G01 & 2.454 & \third{1.809} & 1.996 & \second{1.682} & -- & \best{0.192} \\
    G02 & 2.561 & 2.002 & \third{1.874} & \second{1.848} & -- & \best{0.501} \\
    C001 & 2.473 & \third{1.665} & 2.104 & \second{1.643} & -- & \best{0.241} \\
    C010 & 2.043 & \second{0.656} & \third{1.763} & 2.271 & -- & \best{0.000} \\
    C005 & 2.235 & \second{1.559} & \third{1.938} & 1.979 & -- & \best{0.461} \\
    C014 & 2.383 & \second{1.340} & 1.745 & \third{1.456} & -- & \best{0.718} \\
    T300C010 & 2.238 & \second{0.965} & \third{1.435} & 2.049 & -- & \best{0.047} \\
    T300C021 & 2.359 & \best{0.803} & \third{1.511} & 1.774 & -- & \second{0.951} \\
    \midrule
    Avg. & 2.343 & \second{1.350} & \third{1.796} & 1.838 & -- & \best{0.389} \\
    \bottomrule}
  \setlength{\tabcolsep}{0pt}
  \settowidth{\dimen0}{\begin{tabular}{lcccccc}\tabphysbody\end{tabular}}
  \setlength{\tabcolsep}{\dimexpr(\columnwidth-\dimen0)/14\relax}
  \begin{tabular}{lcccccc}\tabphysbody\end{tabular}
\end{table}

\begin{table}[t]
  \centering
  \caption{Relative error of the half-light radius $R_{50}$ $\downarrow$, measured on deprojected renderings from the unseen near face-on view.}
  \label{tab:r50}
  \scriptsize
  \def\tabphysbody{%
    \toprule
    Scene & 4DGS & FreeGave & MoDec-GS & 4C4D & RetimeGS & Ours \\
    \midrule
    G01 & 0.060 & 0.058 & 0.057 & \third{0.042} & \second{0.031} & \best{0.009} \\
    G02 & 0.131 & 0.130 & 0.082 & \second{0.051} & \third{0.062} & \best{0.019} \\
    C001 & 0.288 & \third{0.280} & 0.323 & 0.307 & \second{0.237} & \best{0.163} \\
    C010 & 0.250 & \best{0.180} & 0.233 & 0.232 & \third{0.218} & \second{0.194} \\
    C005 & \second{0.106} & \best{0.083} & 0.125 & 0.272 & \third{0.109} & 0.122 \\
    C014 & 0.272 & 0.247 & 0.280 & \third{0.214} & \second{0.170} & \best{0.129} \\
    T300C010 & \third{0.505} & 0.506 & 0.514 & \second{0.464} & 0.545 & \best{0.125} \\
    T300C021 & 0.724 & 0.730 & \third{0.721} & \second{0.705} & 0.739 & \best{0.199} \\
    \midrule
    Avg. & 0.292 & \third{0.277} & 0.292 & 0.286 & \second{0.264} & \best{0.120} \\
    \bottomrule}
  \setlength{\tabcolsep}{0pt}
  \settowidth{\dimen0}{\begin{tabular}{lcccccc}\tabphysbody\end{tabular}}
  \setlength{\tabcolsep}{\dimexpr(\columnwidth-\dimen0)/14\relax}
  \begin{tabular}{lcccccc}\tabphysbody\end{tabular}
\end{table}

\paragraph{Protocol.}
Even LPIPS assesses each view independently, and plausible images may arise from primitives that fade and reappear rather than orbit. We therefore export the 3D Gaussian centers and luminosity weights of each method at all 96 frames with its own rendering code, express them in a disk frame fixed by the anchor snapshot, and compare them with the ground-truth particles, which are used only for evaluation. Velocities come from central differences, and all errors use the same luminosity weighting.

\paragraph{Rotation Curves.}
\cref{tab:vphi} reports the relative error of the luminosity-weighted rotation curve $v_\phi(R)$ over 12 radial bins. 4DGS, FreeGave, and MoDec-GS reach $0.85$ to $1.00$, so their Gaussians barely orbit and the apparent rotation comes from changing color and opacity. \method lowers the average error to $0.357$, $43\%$ below the best baseline, and ranks first on six scenes, although our TNG300 disks spin $1.5$ to $2.3\times$ too fast and C005 and G02 too slowly.

\paragraph{Angular Momentum Conservation.}
In a nearly axisymmetric potential, the angular momentum $L_z=Rv_\phi$ of each star is approximately conserved. For every primitive tracked over at least ten frames we measure the relative drift of $L_z$, take the luminosity-weighted median, and compare it in dex with the drift of the ground-truth particles. As shown in \cref{tab:lz}, \method is closest to real stars on seven scenes, averaging $0.389$\,dex, $71\%$ below the $1.350$\,dex of the best baseline. Baseline Gaussians follow paths that violate it, and RetimeGS primitives live for only about four frames.

\begin{figure*}[t]
\centering
\includegraphics[width=\linewidth]{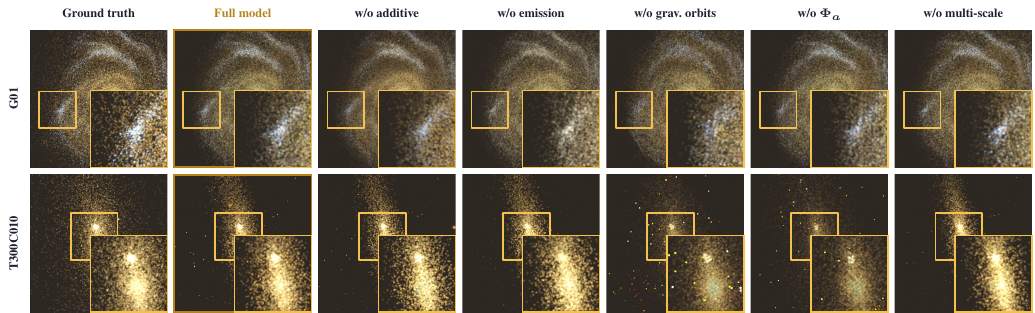}
\caption{Ablations rendered on test views of G01 (top) and T300C010 (bottom). Without additive splatting, front stars dim those behind them. Without evolving emission, the young blue stars fade to grey. Without gravitational orbits or the anchor potential, bright stars drift off the disk and blur the cluster. The insets reveal these details in a young star cluster on the outer arm of G01 and in the core of T300C010.}
\label{fig:ablation}
\end{figure*}
\begin{table*}[t]
  \centering
  \caption{Ablation on the test views. Mask and Full are metrics inside the galaxy foreground mask and on the whole image. RMSE (ckpc$/h$) is the position error of the modeled stars against ground-truth particles, and $v_\phi$ is the relative rotation-curve error used in \cref{tab:vphi}.}
  \label{tab:ablation}
  \footnotesize
  \setlength{\tabcolsep}{3.2pt}
  \resizebox{\linewidth}{!}{%
  \begin{tabular}{l|ccc|ccc|cc|ccc|ccc|cc}
    \toprule
    \multirow{3}{*}{\textbf{Setting}} & \multicolumn{8}{c|}{G01} & \multicolumn{8}{c}{T300C010} \\
    & \multicolumn{3}{c|}{Mask} & \multicolumn{3}{c|}{Full} & \multicolumn{2}{c|}{Physics} & \multicolumn{3}{c|}{Mask} & \multicolumn{3}{c|}{Full} & \multicolumn{2}{c}{Physics} \\
    & PSNR$\uparrow$ & SSIM$\uparrow$ & LPIPS$\downarrow$ & PSNR$\uparrow$ & SSIM$\uparrow$ & LPIPS$\downarrow$ & RMSE$\downarrow$ & $v_\phi$$\downarrow$ & PSNR$\uparrow$ & SSIM$\uparrow$ & LPIPS$\downarrow$ & PSNR$\uparrow$ & SSIM$\uparrow$ & LPIPS$\downarrow$ & RMSE$\downarrow$ & $v_\phi$$\downarrow$ \\
    \midrule
    w/o additive splatting & 19.19 & 0.326 & 0.391 & 25.16 & 0.701 & \best{0.322} & \third{8.61} & 0.185 & 16.66 & 0.168 & \third{0.356} & 23.51 & 0.558 & \third{0.419} & 24.57 & \third{0.740} \\
    w/o evolving emission & \second{20.32} & \third{0.368} & 0.405 & \second{26.33} & \third{0.705} & 0.350 & \best{8.58} & \third{0.125} & \third{17.02} & \second{0.178} & \second{0.337} & \best{23.78} & \third{0.560} & \second{0.404} & 24.58 & 0.751 \\
    \midrule
    w/o gravitational orbits & 19.46 & 0.288 & 0.395 & 25.47 & 0.673 & 0.353 & 8.88 & 0.154 & \second{17.18} & 0.158 & 0.400 & \third{23.65} & 0.553 & 0.465 & \best{23.71} & \second{0.506} \\
    w/o anchor potential & 20.00 & 0.338 & \third{0.379} & 26.02 & 0.692 & 0.342 & 8.92 & 0.232 & \best{17.24} & \third{0.175} & 0.359 & 23.64 & 0.553 & 0.429 & \second{24.00} & \best{0.286} \\
    w/o multi-scale & \third{20.12} & \second{0.372} & \second{0.366} & \third{26.15} & \second{0.708} & \third{0.336} & \second{8.60} & \second{0.122} & 16.64 & 0.165 & 0.427 & 23.59 & \best{0.564} & 0.499 & \third{24.55} & 0.783 \\
    \midrule
    Full model & \best{20.44} & \best{0.387} & \best{0.363} & \best{26.45} & \best{0.712} & \second{0.330} & \second{8.60} & \best{0.121} & 17.01 & \best{0.183} & \best{0.330} & \second{23.74} & \second{0.561} & \best{0.387} & 24.58 & 0.776 \\
    \bottomrule
  \end{tabular}}
\end{table*}
\paragraph{Half-Light Radius.}
The half-light radius $R_{50}$ needs only rendered images, so all methods are compared on equal terms. We deproject every frame of an unseen near face-on view onto the disk plane and find the radius enclosing half of the light over 24 rings. As shown in \cref{tab:r50}, \method reduces the average relative error by $55\%$, from $0.264$ for the best baseline to $0.120$, and is best on six scenes. Because the images are tone-mapped, $R_{50}$ is an apparent radius.

\section{Ablation}
\label{sec:ablation}

We ablate the main components on G01, with both bright and faint particles, and on the sparse T300C010, whose particles are all bright. \cref{tab:ablation} reports image and physical metrics, and Appendix~\ref{sec:suppl_ablation} gives more detailed results.

\paragraph{Optically Thin Rendering and Evolving Emission.}
Replacing additive splatting with alpha compositing causes the largest drop on G01, $1.25$\,dB in masked PSNR, and raises masked LPIPS by $8\%$ on G01, which confirms that an optically thin galaxy should not be rendered with occlusion. When each star keeps its anchor color over time, SSIM falls on both scenes and LPIPS rises by $12\%$ on G01. \cref{fig:ablation} shows that the young blue stars then fade to grey, so evolving emission is needed to follow stellar aging.

\paragraph{Luminosity-Stratified Dynamics.}
Carrying all particles with the shared flow field instead of giving bright stars individual orbits lowers masked SSIM by $26\%$ on G01 and $14\%$ on T300C010 and raises LPIPS on both, and on G01 the rotation-curve error grows from $0.121$ to $0.154$. Keeping only the full-resolution level of the pyramid costs $0.32$ and $0.37$\,dB, and on T300C010 masked LPIPS grows by $29\%$, as misplaced stars in this sparse scene lose gradients.

\paragraph{Gravitational Dynamics.}
Removing the anchor potential $\Phi_a$ and keeping only the learned force $\mathbf{f}_\psi$ lowers masked SSIM by $13\%$ on G01 and $4\%$ on T300C010 and raises LPIPS on both, and on G01 the trajectory and rotation-curve errors grow from $8.60$ to $8.92$\,ckpc$/h$ and from $0.121$ to $0.232$. On T300C010, whose fitted disk spins too fast, both orbit ablations lower $v_\phi$, yet \cref{fig:ablation} shows their bright stars scattering away from the disk.

\section{Conclusion}
\label{sec:conclusion}
We present \method, which reconstructs an evolving galaxy in four dimensions from multi-view images. Stars become self-luminous Gaussians that fall through a gravitational potential, which confines their motion to physically plausible orbits. Luminosity-stratified dynamics trace bright stars along individual orbits and carry faint ones with a shared flow. On eight simulated galaxies, \method renders the sharpest stellar texture among dynamic Gaussian methods and best recovers rotation, angular momentum, and galaxy size. Future work could refine the potential jointly with the images and move toward real observations.

\appendix

\section{Astrophysical Background}
\label{sec:suppl_background}

\begin{figure*}[t]
\centering
\includegraphics[width=\linewidth]{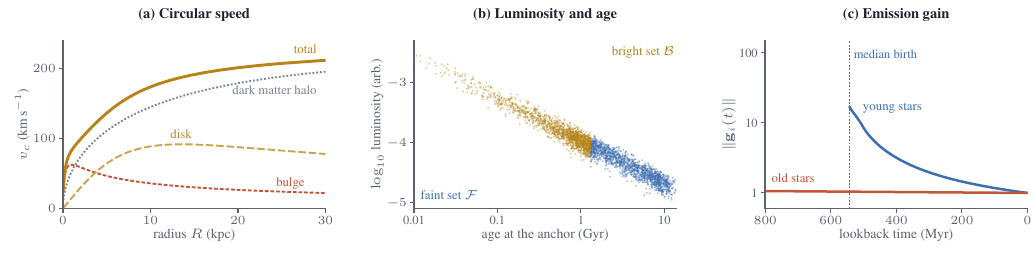}
\caption{Background quantities on G01. (a) Circular speed of the anchor potential $\Phi_a$ fitted in \cref{sec:suppl_dynamics} and of its bulge, disk, and dark matter halo components, which add in quadrature. (b) Luminosity against age at the anchor for 1,500 sampled stars each of the bright set $\mathcal{B}$ and the faint set $\mathcal{F}$. (c) Median emission gain of \cref{sec:suppl_appearance} over lookback time for 400 young stars born during the sequence and 400 old stars, normalized at the anchor.}
\label{fig:background}
\end{figure*}

This section summarizes the astrophysics behind our setting for readers outside astronomy, and Binney and Tremaine~\cite{binney2008galactic} give a complete treatment. \Cref{fig:background} illustrates the main quantities on G01.

\paragraph{Units and Scales.}
Galactic distances are measured in kiloparsecs, and $1$\,kpc is about $3{,}260$ light-years. Cosmological simulations store positions in comoving units of ckpc$/h$, where $h$ is the dimensionless Hubble constant. At the present epoch, which ends every sequence, $1$\,ckpc$/h$ equals $1/h$\,kpc, or about $1.5$\,kpc for the TNG value $h=0.6774$. Times are given in Myr and Gyr, and lookback time counts backward from the anchor. Stellar speeds are quoted in km\,s$^{-1}$, and $1$\,km\,s$^{-1}$ is about $1.02$\,kpc per Gyr. A star moving at $200$\,km\,s$^{-1}$ thus travels about $160$\,kpc during the $0.8$\,Gyr spanned by G01, far more than the size of the disk.

\begin{figure}[t]
\centering
\includegraphics[width=\linewidth]{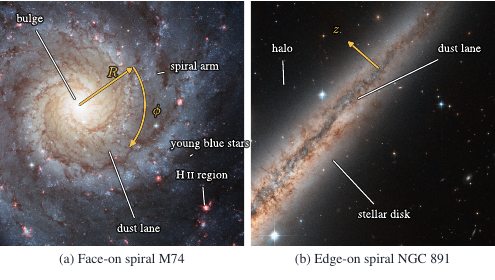}
\caption{Real disk galaxies imaged by Hubble. (a) The face-on spiral M74 with radius $R$ and azimuth $\phi$ in the disk plane. (b) The edge-on spiral NGC 891 with height $z$ above the disk.}
\label{fig:real_galaxies}
\end{figure}

\paragraph{Anatomy of a Disk Galaxy.}
A disk galaxy such as G01 has three main stellar components. The thin rotating disk holds most young stars and the spiral arms, the central bulge is a compact spheroid of mostly old stars, and a sparse stellar halo surrounds both. \Cref{fig:real_galaxies} shows these components in two nearby spirals imaged by Hubble, together with the cylindrical coordinates $(R,\phi,z)$ used in this paper. The M74 image is credited to NASA, ESA, and The Hubble Heritage (STScI/AURA)-ESA/Hubble Collaboration, and the NGC 891 image to ESA/Hubble \& NASA with acknowledgement to Nick Rose, both released under CC BY 4.0. All three sit inside a much larger halo of dark matter, which emits no light yet dominates the mass beyond the inner few kiloparsecs. \Cref{fig:background}a shows these contributions through the circular speed $v_c(R)=\sqrt{R\,\partial\Phi/\partial R}$, the speed of a circular orbit at radius $R$ in the disk plane. The dark matter halo keeps $v_c$ nearly flat at large radii, a hallmark of disk galaxies. Spiral arms are patterns through which stars pass rather than fixed groups of stars, so following an arm in the images does not follow its current stars.

\paragraph{Collisionless Orbits.}
A star is about a hundred-millionth the size of the distance to its nearest neighbors, so stars almost never collide or deflect one another. Each star instead moves in the smooth gravitational potential $\Phi$ of the whole galaxy, which follows from the mass density $\varrho$ through the Poisson equation $\nabla^2\Phi=4\pi G\varrho$ with gravitational constant $G$. Disk stars travel on nearly circular orbits with small radial and vertical oscillations. These oscillations do not repeat with the orbital period, so an orbit traces a rosette rather than a closed curve. Inner stars complete an orbit faster than outer ones, so the disk rotates differentially and any pattern drawn on it winds up over time. Near the Sun an orbit takes about $200$\,Myr~\cite{blandhawthorn2016galaxy}. In G01, a circular orbit takes about $165$\,Myr at $R=3$\,kpc and $480$\,Myr at $R=15$\,kpc, so the sequence covers about five inner orbits but fewer than two outer ones.

\paragraph{Conserved Quantities.}
In a potential that is symmetric around the disk axis and constant in time, a star keeps its energy and its angular momentum about the axis, $L_z=Rv_\phi$, where $v_\phi$ is its azimuthal velocity. Bars and spiral arms break this symmetry slightly and exert small torques, so $L_z$ drifts slowly in real galaxies. For this reason \cref{tab:lz} compares the drift of each method with that of the ground-truth particles rather than with zero. The rotation curve $v_\phi(R)$, the mean azimuthal velocity of stars at radius $R$, traces the mass enclosed within $R$ and is the most common kinematic summary of a disk galaxy.

\paragraph{Stellar Populations.}
A star particle in a cosmological simulation is not a single star. It stands for a population of stars born together, with a mass of about $10^5$ solar masses in TNG50 and up to about $10^7$ in TNG300, and it carries a formation time and a metallicity, the mass fraction of elements heavier than helium. Massive stars are hot, blue, and extremely luminous but live only a few to a few tens of Myr. A young population is therefore bright and blue, and it fades and reddens as its massive members die. Metal-rich populations are also redder at a given age. \Cref{fig:background}b shows the fading on G01, where young particles form most of the bright set $\mathcal{B}$ and old particles most of the faint set $\mathcal{F}$. \Cref{fig:background}c shows the resulting emission gain, which falls by more than an order of magnitude between birth and the anchor for young stars and stays nearly constant for old ones. New particles form throughout the sequence, so part of the light at the anchor is absent from the earliest frames.

\paragraph{Optically Thin Light.}
Because stars cover a negligible fraction of the sky even in a dense disk, light from distant stars reaches the camera past nearer ones. Without dust, an image is the sum of the light of all stars along each line of sight. Dust in real galaxies absorbs and reddens part of this light, and our renderings omit it as described in \cref{sec:suppl_data}. This additive image formation motivates rendering each star by additive splatting in \cref{sec:gravity_gaussians}.

\section{Dataset}
\label{sec:suppl_data}

\paragraph{Simulation Snapshots.}
Each scene is a galaxy taken from a cosmological hydrodynamical simulation, which evolves dark matter, gas, and stars under gravity and baryonic physics and stores the full particle state at discrete times called snapshots. We use the subbox outputs of TNG50-1~\cite{pillepich2019tng50,nelson2019tng50}, TNG100-1, TNG300-1, and Illustris-1, which save a fixed subvolume at a much higher cadence than the full box and therefore resolve galactic motion between frames. For every scene we keep the last 96 subbox snapshots, so the final snapshot serves as the anchor. Each snapshot is recentered on the galaxy when the data are generated. \Cref{tab:scenes} lists the scenes and their particle counts.

\begin{table}[h]
  \centering
  \footnotesize
  \caption{Scenes used in our experiments. $N$ is the number of sampled stellar particles, split into bright and faint sets.}
  \label{tab:scenes}
  \begin{tabular}{llrrr}
    \toprule
    Scene & Source & $N$ & $|\mathcal{B}|$ & $|\mathcal{F}|$ \\
    \midrule
    G01 & TNG50-1 Subbox0 & 110,013 & 56,572 & 53,441 \\
    G02 & TNG50-1 Subbox0 & 110,453 & 45,217 & 65,236 \\
    C001 & TNG100-1 Subbox1 & 75,123 & 75,123 & 0 \\
    C010 & TNG100-1 Subbox0 & 77,379 & 77,379 & 0 \\
    C005 & Illustris-1 Subbox0 & 110,265 & 3,159 & 107,106 \\
    C014 & Illustris-1 Subbox0 & 90,324 & 90,324 & 0 \\
    T300C010 & TNG300-1 Subbox1 & 18,795 & 18,795 & 0 \\
    T300C021 & TNG300-1 Subbox1 & 9,332 & 9,332 & 0 \\
    \bottomrule
  \end{tabular}
\end{table}

\paragraph{Stellar Sampling.}
Candidates are the stars already formed within a scene-scaled radius of $115\,\mathrm{ckpc}/h$. Each star enters the sample with probability $p_i=\min(1,\eta\max_t L_i(t))$, where $\eta$ is chosen so that $\sum_i p_i\approx110{,}000$, through a single Bernoulli draw. The same identities are used in every frame and each star is rendered with luminosity $L_i(t)/p_i$, which keeps the linear flux unbiased and avoids the frame-to-frame flicker of per-snapshot resampling. Stars with $p_i=1$ form the bright set $\mathcal{B}$ of \cref{sec:hybrid_dynamics}, and sampled stars with $p_i<1$ form the faint set $\mathcal{F}$, each standing in for about $1/p_i$ stars. If every formed star is sampled, the scene has no faint stars.

\begin{figure}[t]
\centering
\includegraphics[width=\linewidth]{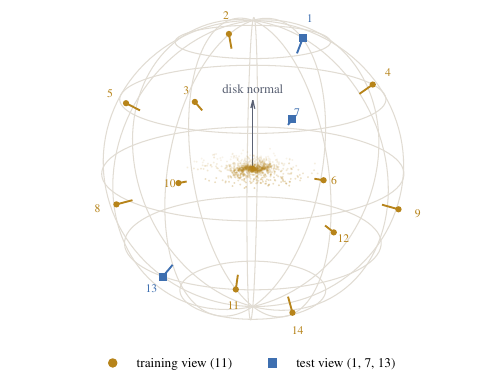}
\caption{Camera layout on G01, drawn with the disk normal pointing up. Fourteen cameras lie on a Fibonacci sphere around the galaxy. Gold circles are training views, blue squares are test views, and each short tick points toward the galaxy center.}
\label{fig:cameras}
\end{figure}

\paragraph{Rendering.}
Fourteen cameras are placed on a Fibonacci sphere, with radius and field of view scaled by the galactic half-light radius. Star colors follow an analytic model of age and metallicity, scaled by the sampled luminosity. Each star is splatted as a point source, radiance is accumulated per band without dust attenuation, and the image is slightly blurred by a Gaussian kernel. Exposure is set per view at the 99.9th percentile with temporal smoothing, and the final image maps each normalized intensity $y$ to $\operatorname{asinh}(7y)/\operatorname{asinh}(7)$. Cameras 1, 7, and 13 are held out for testing. \cref{fig:cameras} shows the camera layout. We exclude the FIRE-2 galaxies m12m and m12f because their snapshot spacing is non-uniform, which makes frame-indexed baselines and our time-integrated dynamics incomparable.

\section{Appearance Model}
\label{sec:suppl_appearance}

\paragraph{Camera Response.}
The camera response is
$\mathcal{T}(\mathbf{E})=\operatorname{asinh}(7\mathbf{E}/X)/\operatorname{asinh}(7)$ with $X=\kappa\,\mathrm{pct}_{99.9}(\mathbf{E})$, where $\kappa$ is an exposure scale and the 99.9th percentile is computed per rendered image under a stop-gradient.

\paragraph{Evolving Emission.}
The anchor emission $\boldsymbol{\ell}_i^a$ is initialized from the age, metallicity, and importance-weighted luminosity $L_i^a/p_i$ of each star. The emission gain in \cref{eq:luminosity} is $\mathbf{g}_i(t)=\mathbf{C}(\mathrm{age}_i(t),Z_i)\,\max(\mathrm{age}_i(t),0.01\,\mathrm{Gyr})^{-0.72}$, normalized by its value at the anchor and linearly interpolated between frames, where $\mathrm{age}_i(t)$ is the age of star $i$ at time $t$, $Z_i$ is its metallicity, and $\mathbf{C}$ is the same age and metallicity color used to initialize $\boldsymbol{\ell}_i^a$. The luminosity correction $\boldsymbol{\rho}_i$ in \cref{eq:luminosity} is linearly interpolated from 13 uniform temporal knots, so its factor stays between $e^{-2}$ and $e^{2}$. Both $\mathbf{g}_i$ and this factor equal one at the anchor, where $\boldsymbol{\ell}_i$ reduces to $\boldsymbol{\ell}_i^a$.

\section{Gravitational Dynamics}
\label{sec:suppl_dynamics}

\paragraph{Anchor Potential.}
The disk frame is centered at the median anchor position, with the disk normal given by the smallest principal axis of particles within $20$\,kpc. The potential combines a Hernquist bulge, a Miyamoto--Nagai disk, and an NFW halo, the three main mass components of a disk galaxy. Disk scale length, scale height, and bulge fraction are estimated from anchor positions only. Dark matter emits no light yet dominates the mass of the outer galaxy, so the halo is derived from $M_\star$. Its mass follows the stellar-to-halo mass relation of Moster \etal~\cite{moster2013galactic}, and its concentration follows Dutton and Macci\`o~\cite{dutton2014cold}. The potential contains no bar or spiral terms, so non-axisymmetric structure is left to the residual force field $\mathbf{f}_\psi$. Bright disk stars start on circular orbits around the disk axis, with the circular speed $v_c(R)$ of $\Phi_a$ at their anchor radius. Disk and halo stars are separated by a two-component exponential mixture over $|z|$ fitted within each radial bin, and stars whose disk posterior is below $0.5$ are labeled as halo. Halo stars start on circular orbits in randomly oriented planes. In thick-disk scenes (C014, T300C010, and T300C021), all bright stars are treated as disk stars.

\paragraph{Residual Force Basis.}
Some fields in \cref{eq:force_basis} are symmetric around the disk axis and adjust the overall pull, while others have two arms ($m=2$) and represent bars and spirals. Each non-uniform field is $\mathbf{F}_q=-\nabla\varphi_q$ with $\varphi_q=\exp[-\tfrac12((R-r_q)/\sigma_q)^2-\tfrac12(z/h_q)^2]A_q(\phi)$. The 36 axisymmetric terms ($A_q=1$) use 12 log-spaced radii in $[0.3,70]$\,kpc and heights $h_q\in\{1,4,20\}$\,kpc. The 16 $m=2$ terms use the middle 8 radii with $A_q\in\{\cos2\phi,\sin2\phi\}$. Three uniform accelerations complete the basis and absorb the independent recentering of the benchmark snapshots, giving 55 fields in total. The sum in \cref{eq:force_basis} is scaled by $10^3$. Coefficients are zero-initialized and defined on 13 uniform temporal knots, which gives 715 shared coefficients.

\paragraph{Leapfrog Integrator.}
We use a kick--drift--kick leapfrog scheme with substeps of at most $1$\,Myr, run in double precision. The scheme alternates small velocity and position updates and conserves energy well over many orbital periods, which makes it a standard choice for stellar dynamics. With step $\delta t$, negative when integrating backward from the anchor, one step updates velocity $\mathbf{v}$ and position $\mathbf{x}$ as
\begin{equation}
\begin{aligned}
\mathbf{v}_{n+\frac12}&=\mathbf{v}_n+\tfrac{\delta t}{2}\,\mathbf{a}_\psi(\mathbf{x}_n,t_n),\\
\mathbf{x}_{n+1}&=\mathbf{x}_n+\delta t\,\mathbf{v}_{n+\frac12},\\
\mathbf{v}_{n+1}&=\mathbf{v}_{n+\frac12}+\tfrac{\delta t}{2}\,\mathbf{a}_\psi(\mathbf{x}_{n+1},t_{n+1}),
\end{aligned}
\label{eq:leapfrog}
\end{equation}
with $\mathbf{a}_\psi$ from \cref{eq:dynamics}. Within each frame interval, the residual force is evaluated at the starting position and interpolated in time, while gravity is evaluated throughout the integration. Gradients follow this discrete map exactly. Hermite interpolation of frame positions and velocities yields $C^1$ motion between frames. Gradients of each frame-interval map are computed in forward mode inside a CUDA kernel and applied as vector--Jacobian products, with gradient checkpointing per interval. On our scenes, the anchor position error is $3.4\times10^{-5}\,\mathrm{ckpc}/h$, and finite-difference checks of directional gradients agree to a relative error of $6\times10^{-4}$.

\section{Blended Orbital Segments}
\label{sec:suppl_segments}

\begin{figure}[t]
\centering
\includegraphics[width=\linewidth]{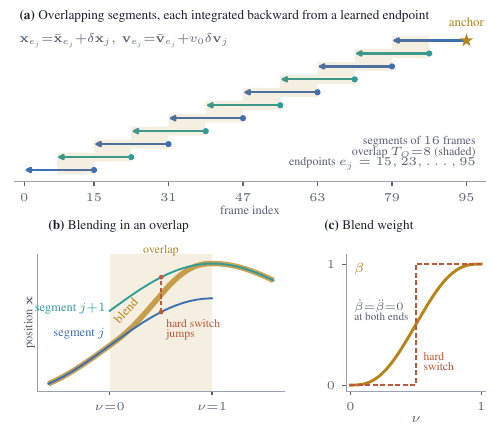}
\caption{Blended orbital segments. (a) Overlapping segments integrated backward from learned endpoints. (b) Blending two segments within an overlap versus a hard switch. (c) Quintic blend weight $\beta$ and the step weight of a hard switch.}
\label{fig:segments}
\end{figure}

The main text describes each bright orbit as one backward integration from the anchor (\cref{sec:gravity_gaussians}). This section describes how we keep these orbits optimizable over long sequences in practice. A single backward integration from the anchor ties early frames to accumulated state errors. Short independent integrations loosen this coupling, but switching abruptly between them produces visible jumps in position and velocity. We instead optimize overlapping segments and render one continuous trajectory. As illustrated in \cref{fig:segments}a, each bright orbit is split into 16-frame segments that overlap by 8 frames. Each segment is integrated backward from a learned endpoint state, and only the last one reaches the anchor. Within an overlap the trajectory passes smoothly from one segment to the next rather than jumping between them (\cref{fig:segments}b and c).

\paragraph{Local Orbit Integration.}
Each bright-particle segment $j$ is integrated backward from endpoint $e_j$, starting from
\begin{equation}
\begin{aligned}
\mathbf{x}_{i,e_j}&=\bar{\mathbf{x}}_{i,e_j}
+\delta\mathbf{x}_{i,j},\\
\mathbf{v}_{i,e_j}&=\bar{\mathbf{v}}_{i,e_j}
+v_0\,\delta\mathbf{v}_{i,j},
\end{aligned}
\label{eq:shooting_states}
\end{equation}
where bars denote analytic seeds and $v_0$ corresponds to $40$ km/s in integration units. Endpoint offsets are learned from images, with the anchor position held fixed. For the 96-frame sequence, segments span 16 frame intervals with an overlap of 8, giving endpoints $15,23,\ldots,95$ and starts $\max(0,e_j-16)$. All segments share the same force field and differentiable integrator.

\paragraph{Partition-of-Unity Blending.}
Within the overlap of adjacent segments $j$ and $j+1$, let $\nu\in[0,1]$ denote the normalized overlap time. We blend the two trajectories as
\begin{equation}
\begin{aligned}
\beta(\nu)&=6\nu^5-15\nu^4+10\nu^3,\\
\mathbf{x}_i(t)&=(1-\beta)\,\mathbf{x}_i^{(j)}(t)
+\beta\,\mathbf{x}_i^{(j+1)}(t).
\end{aligned}
\label{eq:trajectory_assembly}
\end{equation}
The velocity is the exact derivative of this blend, including $\dot\beta(\mathbf{x}_i^{(j+1)}-\mathbf{x}_i^{(j)})$. Because the weight has vanishing first and second derivatives at both ends, the assembly introduces no jumps in position or velocity. With the Hermite interpolants, each bright trajectory is $C^1$, and the anchor, covered only by its own segment, keeps its position exactly.

Continuity, however, does not by itself ensure dynamical consistency. Let $\Delta\mathbf{x}=\mathbf{x}^{(j+1)}-\mathbf{x}^{(j)}$ and $\Delta\mathbf{v}=\mathbf{v}^{(j+1)}-\mathbf{v}^{(j)}$. If both segments satisfy \cref{eq:dynamics}, then the blended trajectory obeys
\begin{equation}
\ddot{\mathbf{x}}-\mathbf{a}_\psi(\mathbf{x},t)
=2\dot\beta\,\Delta\mathbf{v}+\ddot\beta\,\Delta\mathbf{x}
+\mathcal{O}\!\left(\beta(1-\beta)\|\nabla^2\mathbf{a}_\psi\|\,\|\Delta\mathbf{x}\|^2\right),
\label{eq:blend_force}
\end{equation}
up to integration and interpolation error. Non-Newtonian forces thus arise only where segments disagree and vanish once they agree. They also weaken as the overlap length $T_O$ grows, since $\max|\dot\beta|=15/(8T_O)$ and $\max|\ddot\beta|=10/(\sqrt3\,T_O^2)$. Classical multiple shooting amounts to a step weight, which concentrates the same disagreement into an impulse at each node. We therefore treat the result as a physically guided approximation rather than an exact Newtonian solution of the dynamics.

\begin{table}[t]
 \centering
 \caption{Segmented integration on G01 and T300C010. Single segment integrates each bright orbit from the anchor in one piece, and hard switching replaces the blending weight with a step. Metrics are masked, and RMSE is the 3D position error of the modeled stars in ckpc$/h$, used only for diagnosis.}
 \label{tab:segments}
 \footnotesize
 \setlength{\tabcolsep}{2.5pt}
 \resizebox{\linewidth}{!}{%
 \begin{tabular}{l|cccc|cccc}
 \toprule
 \multirow{2}{*}{\textbf{Setting}} & \multicolumn{4}{c|}{G01} & \multicolumn{4}{c}{T300C010} \\
 & PSNR$\uparrow$ & SSIM$\uparrow$ & LPIPS$\downarrow$ & RMSE$\downarrow$ & PSNR$\uparrow$ & SSIM$\uparrow$ & LPIPS$\downarrow$ & RMSE$\downarrow$ \\
 \midrule
 Full model & 20.44 & 0.387 & 0.363 & 8.60 & 17.01 & 0.183 & 0.330 & 24.58 \\
 Single segment & 19.88 & 0.319 & 0.366 & 11.10 & 16.93 & 0.157 & 0.364 & 26.53 \\
 Hard switching & 20.43 & 0.385 & 0.364 & 8.60 & 16.97 & 0.180 & 0.335 & 24.60 \\
 \bottomrule
 \end{tabular}}
\end{table}

Integrating each bright orbit as a single segment from the anchor degrades rendering on both scenes and raises the trajectory error from $8.60$ to $11.10$\,ckpc$/h$ on G01 and from $24.58$ to $26.53$\,ckpc$/h$ on T300C010 (\cref{tab:segments}), which supports the segmented formulation. Hard switching between segments changes every metric only marginally, so the blending mainly provides continuous motion rather than better image fits.

\begin{table*}[t]
  \centering
  \caption{Complete ablation on G01 and T300C010. Mask and Full denote metrics within the galaxy foreground mask and on the whole image. RMSE is the 3D position error of modeled stars against ground-truth particles in ckpc$/h$ and is used only for diagnosis. The full model is listed first.}
  \label{tab:ablation_full}
  \footnotesize
  \setlength{\tabcolsep}{2.0pt}
  \begin{tabular}{l|ccc|ccc|c|ccc|ccc|c}
    \toprule
    \multirow{3}{*}{\textbf{Setting}} & \multicolumn{7}{c|}{G01} & \multicolumn{7}{c}{T300C010} \\
    & \multicolumn{3}{c|}{Mask} & \multicolumn{3}{c|}{Full} & & \multicolumn{3}{c|}{Mask} & \multicolumn{3}{c|}{Full} & \\
    & PSNR$\uparrow$ & SSIM$\uparrow$ & LPIPS$\downarrow$ & PSNR$\uparrow$ & SSIM$\uparrow$ & LPIPS$\downarrow$ & RMSE$\downarrow$ & PSNR$\uparrow$ & SSIM$\uparrow$ & LPIPS$\downarrow$ & PSNR$\uparrow$ & SSIM$\uparrow$ & LPIPS$\downarrow$ & RMSE$\downarrow$ \\
    \midrule
    Full model & 20.44 & 0.387 & 0.363 & 26.45 & 0.712 & 0.330 & 8.60 & 17.01 & 0.183 & 0.330 & 23.74 & 0.561 & 0.387 & 24.58 \\
    \midrule
    w/o additive splatting & 19.19 & 0.326 & 0.391 & 25.16 & 0.701 & 0.322 & 8.61 & 16.66 & 0.168 & 0.356 & 23.51 & 0.558 & 0.419 & 24.57 \\
    w/o evolving emission & 20.32 & 0.368 & 0.405 & 26.33 & 0.705 & 0.350 & 8.58 & 17.02 & 0.178 & 0.337 & 23.78 & 0.560 & 0.404 & 24.58 \\
    w/o residual force & 20.44 & 0.386 & 0.362 & 26.45 & 0.711 & 0.330 & 8.63 & 16.97 & 0.182 & 0.331 & 23.72 & 0.561 & 0.385 & 24.57 \\
    \midrule
    w/o gravitational orbits & 19.46 & 0.288 & 0.395 & 25.47 & 0.673 & 0.353 & 8.88 & 17.18 & 0.158 & 0.400 & 23.65 & 0.553 & 0.465 & 23.71 \\
    w/o anchor potential & 20.00 & 0.338 & 0.379 & 26.02 & 0.692 & 0.342 & 8.92 & 17.24 & 0.175 & 0.359 & 23.64 & 0.553 & 0.429 & 24.00 \\
    w/o multi-scale & 20.12 & 0.372 & 0.366 & 26.15 & 0.708 & 0.336 & 8.60 & 16.64 & 0.165 & 0.427 & 23.59 & 0.564 & 0.499 & 24.55 \\
        \bottomrule
  \end{tabular}
\end{table*}

\begin{figure*}[t]
\centering
\includegraphics[width=\linewidth]{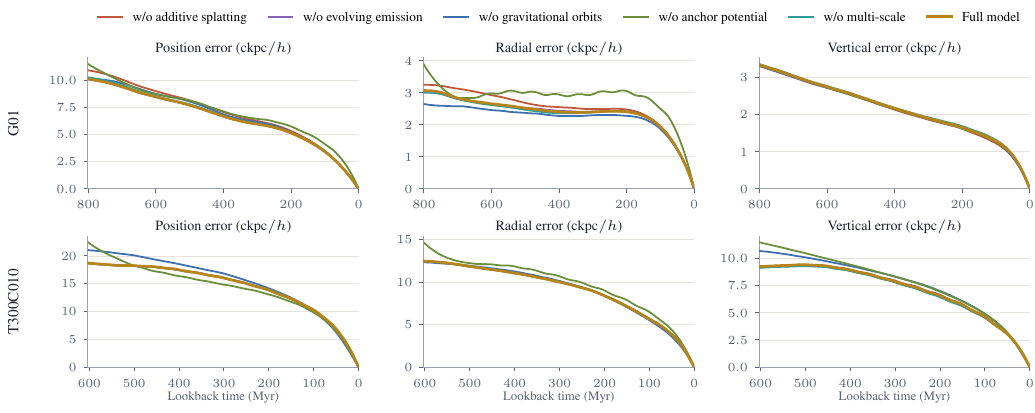}
\caption{Trajectory error of all modeled stars against ground-truth particles over lookback time for each ablation variant on G01 (top) and T300C010 (bottom). Columns show the 3D position error and its radial and vertical components in ckpc$/h$.}
\label{fig:traj_errors}
\end{figure*}

\section{Faint-Star Flow}
\label{sec:suppl_flow}

\paragraph{Flow Estimation.}
Optical flow is computed with DIS~\cite{kroeger2016dis} on $512^2$ grayscale training views in both directions. At sampled disk-plane locations, camera projection Jacobians map each three-dimensional basis velocity to a pixel displacement, and the coefficients are recovered by robust least squares weighted by visibility, brightness, and forward--backward consistency. The disk-plane sampling grid has 32 log-spaced radii and 96 azimuths. The velocity basis has 99 coefficients per frame, comprising 9 rotational, 72 radial/tangential $m\in\{2,4\}$, 9 vertical, and 9 radial components. We solve ridge-regularized weighted least squares with four reweighting iterations and smooth the coefficients temporally with a Gaussian of standard deviation 2 frames.

\paragraph{Bounded Corrections.}
The deformation network in \cref{eq:faint_residual} outputs $\mathcal{D}_\theta=b_{\max}\tanh(\widetilde{\mathcal{D}}_\theta)$, where $\widetilde{\mathcal{D}}_\theta$ is the raw MLP output and the componentwise bound is $b_{\max}=5\,\mathrm{ckpc}/h$ by default and $40\,\mathrm{ckpc}/h$ on C005. The shared phase correction is $\alpha(\hat t)=\tfrac12(\hat t-1)\tanh\rho_{\mathrm{rot}}(\hat t)$, with $\rho_{\mathrm{rot}}$ linearly interpolated from 13 uniform temporal knots, so that $|\alpha|\le0.5$\,rad and $\alpha$ vanishes at the anchor. The neural residual uses a multi-resolution HexPlane with a zero-initialized output head, and its learning rates are rescaled by the scene normalization factor so that they match the effective rates of 4DGS~\cite{Wu_2024_CVPR}.

\begin{figure*}[t]
\centering
\includegraphics[width=\linewidth]{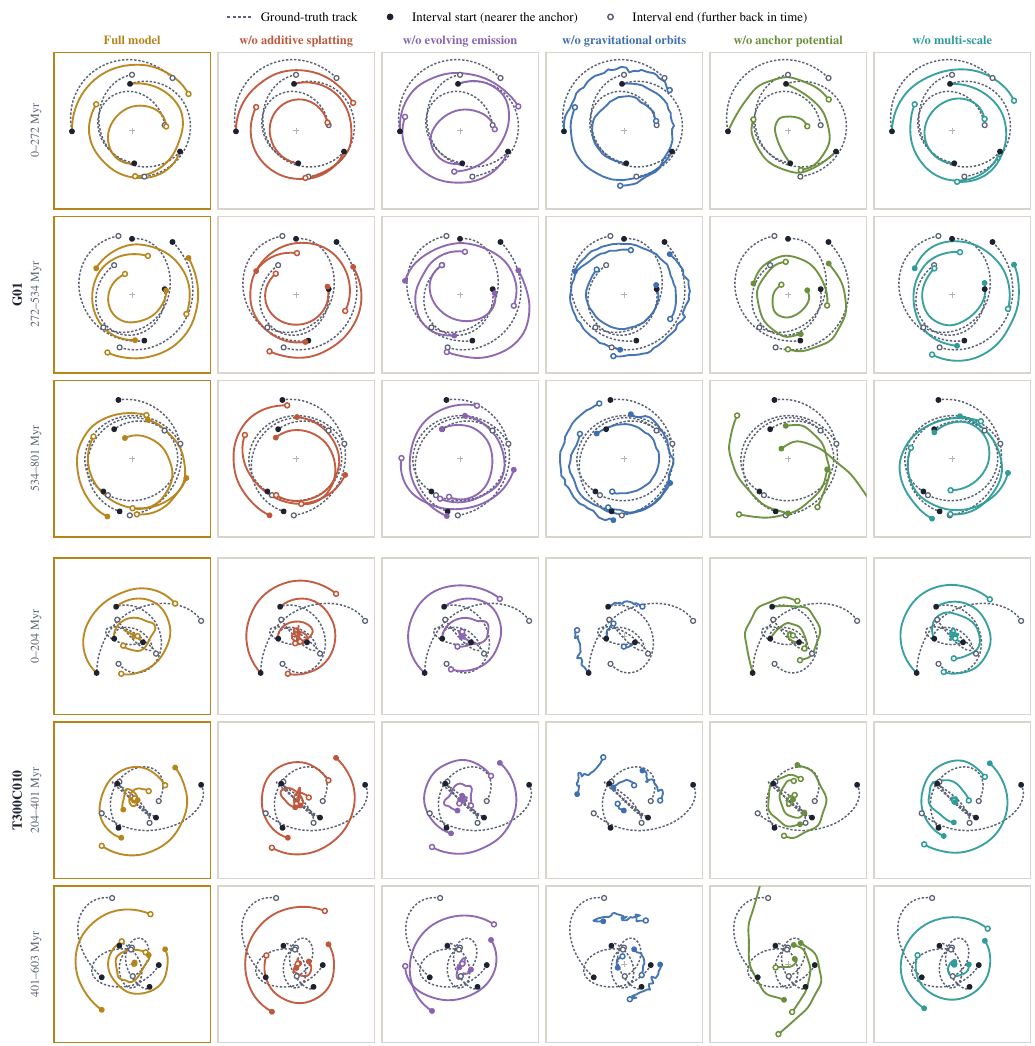}
\caption{Disk-plane tracks of four bright stars for each ablation variant on G01 (top) and T300C010 (bottom), with the full model first and the ablations in the order of \cref{tab:ablation}. Each row covers one third of the sequence, counted back in lookback time from the anchor. Dashed gray lines are ground-truth particle tracks, and solid lines are the modeled tracks in the color of the column title. Filled dots mark the start of each interval and open circles its end, in black or gray for the ground truth and in the variant color for the model. The cross marks the galaxy center. Ground-truth particles are used for evaluation only.}
\label{fig:ablation_full}
\end{figure*}

\section{Implementation Details}
\label{sec:suppl_impl}

\paragraph{Gaussians and Rasterizer.}
Each Gaussian keeps a fixed one-to-one correspondence with a sampled stellar particle, and Gaussian centers have no free parameters and are always produced by the dynamics module. Rotations are frozen and only the DC color component is used. Splat radii are $r_i=r_{\max}\tanh(\tilde r_i/r_{\max})$ with $r_{\max}=0.07\,\mathrm{ckpc}/h$, and are learned only during the anchor-only stage. Opacities are initialized to $0.08$. The additive rasterizer skips contributions below $1/1024$ and applies no output blur.

\paragraph{Losses.}
The pixel weight in \cref{eq:pyramid} is $w(\mathbf{u})=1+20\sqrt{\max_{\mathrm{rgb}}I(\mathbf{u})}$, where the constant keeps a unit weight on the dark sky, and at the coarse levels the pixel weights are average-pooled together with the images. In \cref{eq:regularizer} we use $\lambda_{\psi}=10^{-4}$ with $\gamma_{\psi}=1$ and $\lambda_{\boldsymbol{\rho}}=10^{-5}$ with $\gamma_{\boldsymbol{\rho}}=0.2$. The phase curve $\rho_{\mathrm{rot}}$ of the faint-star rotation is regularized in the same form, with $\lambda=10^{-4}$ and $\gamma=1$.

\paragraph{Optimization.}
The anchor-only stage uses images at the anchor time only and keeps positions static. In the joint stage, every base learning rate is multiplied by
$\min(1,\tau/400)\,\bigl(0.1+0.45(1+\cos\pi\varpi)\bigr)$ with $\varpi=\operatorname{clip}\bigl((\tau-400)/23600,0,1\bigr)$ at iteration $\tau$, which gives a 400-iteration linear warmup followed by cosine decay to a 10\% floor at iteration 24,000. Base learning rates are $10^{-3}$ for color, opacity, and force coefficients, $0.025$ for shooting position offsets, and $0.01$ for shooting velocity offsets and luminosity corrections. Dynamics, velocity, and neural-field gradients are clipped to norm $0.1$. Image resolution grows from $256^2$ to the full $512^2$, reached at iteration 801. The anchor-to-past curriculum samples frames uniformly from the last $K=\min(96,1+\lfloor(\tau-1)/8\rfloor)$ frames ending at the anchor, so all 96 frames are covered from iteration 761. We report iteration 32,000 without test-view checkpoint selection. Training takes 6--20 hours per scene on one GPU, depending on the particle count.

\paragraph{Hyperparameter Selection.}
\label{sec:suppl_selection}
Hyperparameters and the few scene-dependent settings are selected on a held-out training camera, trained for 8,000 iterations, using a rule fixed in advance. That camera is excluded from flow estimation during validation. Test cameras are never used for selection. The scene-dependent settings are the exposure scale $\kappa\in\{1.0,1.75\}$, whether a separate halo population is modeled for seeding, and the faint residual bound $b_{\max}$, which is $5\,\mathrm{ckpc}/h$ except $40\,\mathrm{ckpc}/h$ for C005.

\section{Additional Ablations}
\label{sec:suppl_ablation}

\cref{tab:ablation_full} lists every ablation on the two ablation scenes, with masked and full-image metrics and the trajectory error of the modeled stars. Ground-truth particles enter only this diagnostic and never training or selection. Freezing the residual force $\mathbf{f}_\psi$ changes PSNR by less than $0.1$\,dB and SSIM and LPIPS by less than $0.01$. Each variant is trained with a single seed, so differences of this size should not be read as meaningful. \cref{fig:traj_errors} follows the trajectory error of each variant over the sequence. The error is zero at the anchor and grows back in time for every variant, and removing the anchor potential raises the radial error over most of the sequence on both scenes. \cref{fig:ablation_full} plots the disk-plane tracks of sample bright stars for every variant.

\section{Limitations}
\label{sec:suppl_limitations}

\paragraph{Single-View Observations.}
A telescope observes each galaxy from a single direction, and galactic rotation is far too slow to change the image over a human lifetime. A real observation therefore provides one view at one instant, whereas \method, like other dynamic scene reconstruction methods, requires multiple calibrated views over time. For this reason our experiments use simulated galaxies rendered into 14 cameras. Applying the method to real data requires priors that supply the missing views and epochs. Generative models that synthesize multi-view videos from sparse inputs~\cite{wu2025cat4d} are one option, and our gravitational dynamics could then constrain the generated views toward consistent orbital motion.

\paragraph{Evaluation.}
Ground-truth three-dimensional positions and velocities of individual stars are unavailable for real galaxies, so both the image and the physical metrics are computed on simulated scenes. The evaluation is thus bounded by how closely TNG and Illustris galaxies, with their finite particle resolution and subgrid physics, resemble real ones. Real observations offer only indirect checks, such as spectroscopic line-of-sight velocities, which constrain a single velocity component and do not yield full trajectories.

\paragraph{Dust.}
Our renderer treats stars as optically thin emitters. In real galaxies, interstellar dust absorbs and scatters starlight and dims and reddens regions of active star formation~\cite{calzetti2000dust}. Dust lanes violate the additive emission model and may be explained as missing stars or altered luminosities. Real data would require wavelength-dependent attenuation along each ray.

\paragraph{Noise and Resolution.}
Our simulated images are noise-free, and stars span a few pixels each. Real deep-sky images contain sensor noise from photon statistics, dark current, and readout~\cite{liu2026denoising}, and the limited angular resolution of telescopes blends many stars into each pixel~\cite{liu2026fluxflow}. Faint stars, which carry much of the flow in our model, are affected most. Physics-based denoising~\cite{liu2026denoising} and flux-conserving super-resolution~\cite{liu2026fluxflow} could serve as preprocessing, and modeling such noise and degradation inside the rendering loss is a natural extension.

{
    \small
    \bibliographystyle{ieeenat_fullname}
    \bibliography{main,cosmos4d-method,cosmos4d-related}

@InProceedings{Cao_2023_CVPR,
  author = {Cao, Ang and Johnson, Justin},
  title = {{HexPlane}: A Fast Representation for Dynamic Scenes},
  booktitle = {Proceedings of the IEEE/CVF Conference on Computer Vision and Pattern Recognition},
  year = {2023},
  pages = {130--141}
}

@InProceedings{Wu_2024_CVPR,
  author = {Wu, Guanjun and Yi, Taoran and Fang, Jiemin and Xie, Lingxi and Zhang, Xiaopeng and Wei, Wei and Liu, Wenyu and Tian, Qi and Wang, Xinggang},
  title = {{4D} Gaussian Splatting for Real-Time Dynamic Scene Rendering},
  booktitle = {Proceedings of the IEEE/CVF Conference on Computer Vision and Pattern Recognition},
  year = {2024},
  pages = {20310--20320}
}

@article{pillepich2019tng50,
  author = {Pillepich, Annalisa and Nelson, Dylan and Springel, Volker and Pakmor, R{\"u}diger and Torrey, Paul and Weinberger, Rainer and Vogelsberger, Mark and Marinacci, Federico and Genel, Shy and van der Wel, Arjen and Hernquist, Lars},
  title = {First results from the {TNG50} simulation: the evolution of stellar and gaseous discs across cosmic time},
  journal = {Monthly Notices of the Royal Astronomical Society},
  volume = {490},
  number = {3},
  pages = {3196--3233},
  year = {2019}
}

@article{nelson2019tng50,
  author = {Nelson, Dylan and Pillepich, Annalisa and Springel, Volker and Pakmor, R{\"u}diger and Weinberger, Rainer and Genel, Shy and Torrey, Paul and Vogelsberger, Mark and Marinacci, Federico and Hernquist, Lars},
  title = {First results from the {TNG50} simulation: galactic outflows driven by supernovae and black hole feedback},
  journal = {Monthly Notices of the Royal Astronomical Society},
  volume = {490},
  number = {3},
  pages = {3234--3261},
  year = {2019}
}

@book{binney2008galactic,
  author = {Binney, James and Tremaine, Scott},
  title = {Galactic Dynamics},
  edition = {Second},
  publisher = {Princeton University Press},
  year = {2008}
}

@InProceedings{li2025freegave,
    author    = {Li, Jinxi and Song, Ziyang and Zhou, Siyuan and Yang, Bo},
    title     = {FreeGave: 3D Physics Learning from Dynamic Videos by Gaussian Velocity},
    booktitle = {Proceedings of the IEEE/CVF Conference on Computer Vision and Pattern Recognition},
    year      = {2025},
    pages     = {12433--12443}
}

@InProceedings{kwak2025modecgs,
    author    = {Kwak, Sangwoon and Kim, Joonsoo and Jeong, Jun Young and Cheong, Won-Sik and Oh, Jihyong and Kim, Munchurl},
    title     = {{MoDec-GS}: Global-to-Local Motion Decomposition and Temporal Interval Adjustment for Compact Dynamic {3D} Gaussian Splatting},
    booktitle = {Proceedings of the IEEE/CVF Conference on Computer Vision and Pattern Recognition},
    year      = {2025},
    pages     = {11338--11348}
}

@InProceedings{zhou2026fourc4d,
    author    = {Zhou, Junsheng and Yang, Zhifan and Han, Liang and Zhang, Wenyuan and Shi, Kanle and Xu, Shenkun and Liu, Yu-Shen},
    title     = {{4C4D}: 4 Camera {4D} Gaussian Splatting},
    booktitle = {Proceedings of the IEEE/CVF Conference on Computer Vision and Pattern Recognition},
    year      = {2026},
    pages     = {11829--11839}
}

@InProceedings{wang2026retimegs,
    author    = {Wang, Xuezhen and Ma, Li and Shen, Yulin and Wang, Zeyu and Sander, Pedro V.},
    title     = {{RetimeGS}: Continuous-Time Reconstruction of {4D} Gaussian Splatting},
    booktitle = {Proceedings of the IEEE/CVF Conference on Computer Vision and Pattern Recognition},
    year      = {2026},
    pages     = {7340--7350}
}

@article{lupton2004rgb,
  author = {Lupton, Robert and Blanton, Michael R. and Fekete, George and Hogg, David W. and O'Mullane, Wil and Szalay, Alex and Wherry, Nicholas},
  title = {Preparing Red-Green-Blue Images from {CCD} Data},
  journal = {Publications of the Astronomical Society of the Pacific},
  volume = {116},
  number = {816},
  pages = {133--137},
  year = {2004}
}

@article{freeman1970disks,
  author = {Freeman, Kenneth C.},
  title = {On the Disks of Spiral and {S0} Galaxies},
  journal = {The Astrophysical Journal},
  volume = {160},
  pages = {811--830},
  year = {1970}
}

@article{hernquist1990bulge,
  author = {Hernquist, Lars},
  title = {An Analytical Model for Spherical Galaxies and Bulges},
  journal = {The Astrophysical Journal},
  volume = {356},
  pages = {359--364},
  year = {1990}
}

@article{miyamoto1975disk,
  author = {Miyamoto, Masanori and Nagai, Ryuzaburo},
  title = {Three-Dimensional Models for the Distribution of Mass in Galaxies},
  journal = {Publications of the Astronomical Society of Japan},
  volume = {27},
  pages = {533--543},
  year = {1975}
}

@article{navarro1997nfw,
  author = {Navarro, Julio F. and Frenk, Carlos S. and White, Simon D. M.},
  title = {A Universal Density Profile from Hierarchical Clustering},
  journal = {The Astrophysical Journal},
  volume = {490},
  number = {2},
  pages = {493--508},
  year = {1997}
}

@article{sellwood2014secular,
  author = {Sellwood, J. A.},
  title = {Secular Evolution in Disk Galaxies},
  journal = {Reviews of Modern Physics},
  volume = {86},
  number = {1},
  pages = {1--46},
  year = {2014}
}

@article{wang2004ssim,
  title   = {Image Quality Assessment: From Error Visibility to Structural Similarity},
  author  = {Wang, Zhou and Bovik, Alan C. and Sheikh, Hamid R. and Simoncelli, Eero P.},
  journal = {IEEE Transactions on Image Processing},
  volume  = {13},
  number  = {4},
  pages   = {600--612},
  year    = {2004}
}

@article{moster2013galactic,
  author = {Moster, Benjamin P. and Naab, Thorsten and White, Simon D. M.},
  title = {Galactic star formation and accretion histories from matching galaxies to dark matter haloes},
  journal = {Monthly Notices of the Royal Astronomical Society},
  volume = {428},
  number = {4},
  pages = {3121--3138},
  year = {2013}
}

@article{dutton2014cold,
  author = {Dutton, Aaron A. and Macci{\`o}, Andrea V.},
  title = {Cold dark matter haloes in the {Planck} era: evolution of structural parameters for {Einasto} and {NFW} profiles},
  journal = {Monthly Notices of the Royal Astronomical Society},
  volume = {441},
  number = {4},
  pages = {3359--3374},
  year = {2014}
}

@article{pillepich2018tng,
  author = {Pillepich, Annalisa and Nelson, Dylan and Hernquist, Lars and Springel, Volker and Pakmor, R{\"u}diger and Torrey, Paul and Weinberger, Rainer and Genel, Shy and Naiman, Jill P. and Marinacci, Federico and Vogelsberger, Mark},
  title = {First results from the {IllustrisTNG} simulations: the stellar mass content of groups and clusters of galaxies},
  journal = {Monthly Notices of the Royal Astronomical Society},
  volume = {475},
  number = {1},
  pages = {648--675},
  year = {2018}
}

@article{springel2018tng,
  author = {Springel, Volker and Pakmor, R{\"u}diger and Pillepich, Annalisa and Weinberger, Rainer and Nelson, Dylan and Hernquist, Lars and Vogelsberger, Mark and Genel, Shy and Torrey, Paul and Marinacci, Federico and Naiman, Jill},
  title = {First results from the {IllustrisTNG} simulations: matter and galaxy clustering},
  journal = {Monthly Notices of the Royal Astronomical Society},
  volume = {475},
  number = {1},
  pages = {676--698},
  year = {2018}
}

@article{nelson2018tng,
  author = {Nelson, Dylan and Pillepich, Annalisa and Springel, Volker and Weinberger, Rainer and Hernquist, Lars and Pakmor, R{\"u}diger and Genel, Shy and Torrey, Paul and Vogelsberger, Mark and Kauffmann, Guinevere and Marinacci, Federico and Naiman, Jill},
  title = {First results from the {IllustrisTNG} simulations: the galaxy colour bimodality},
  journal = {Monthly Notices of the Royal Astronomical Society},
  volume = {475},
  number = {1},
  pages = {624--647},
  year = {2018}
}

@article{vogelsberger2014illustris,
  author = {Vogelsberger, Mark and Genel, Shy and Springel, Volker and Torrey, Paul and Sijacki, Debora and Xu, Dandan and Snyder, Greg and Nelson, Dylan and Hernquist, Lars},
  title = {Introducing the {Illustris} Project: simulating the coevolution of dark and visible matter in the {Universe}},
  journal = {Monthly Notices of the Royal Astronomical Society},
  volume = {444},
  number = {2},
  pages = {1518--1547},
  year = {2014}
}

@article{genel2014illustris,
  author = {Genel, Shy and Vogelsberger, Mark and Springel, Volker and Sijacki, Debora and Nelson, Dylan and Snyder, Greg and Rodriguez-Gomez, Vicente and Torrey, Paul and Hernquist, Lars},
  title = {Introducing the {Illustris} project: the evolution of galaxy populations across cosmic time},
  journal = {Monthly Notices of the Royal Astronomical Society},
  volume = {445},
  number = {1},
  pages = {175--200},
  year = {2014}
}

@inproceedings{kroeger2016dis,
  author = {Kroeger, Till and Timofte, Radu and Dai, Dengxin and Van Gool, Luc},
  title = {Fast Optical Flow Using Dense Inverse Search},
  booktitle = {Computer Vision -- {ECCV} 2016},
  year = {2016},
  pages = {471--488}
}

@inproceedings{zhang2018lpips,
  author = {Zhang, Richard and Isola, Phillip and Efros, Alexei A. and Shechtman, Eli and Wang, Oliver},
  title = {The Unreasonable Effectiveness of Deep Features as a Perceptual Metric},
  booktitle = {Proceedings of the IEEE Conference on Computer Vision and Pattern Recognition},
  year = {2018},
  pages = {586--595}
}

@inproceedings{blau2018perception,
  author = {Blau, Yochai and Michaeli, Tomer},
  title = {The Perception-Distortion Tradeoff},
  booktitle = {Proceedings of the IEEE Conference on Computer Vision and Pattern Recognition},
  year = {2018},
  pages = {6228--6237}
}

@inproceedings{wu2025cat4d,
  author = {Wu, Rundi and Gao, Ruiqi and Poole, Ben and Trevithick, Alex and Zheng, Changxi and Barron, Jonathan T. and Holynski, Aleksander},
  title = {{CAT4D}: Create Anything in {4D} with Multi-View Video Diffusion Models},
  booktitle = {Proceedings of the IEEE/CVF Conference on Computer Vision and Pattern Recognition (CVPR)},
  year = {2025},
  pages = {26057--26068}
}

@article{calzetti2000dust,
  author = {Calzetti, Daniela and Armus, Lee and Bohlin, Ralph C. and Kinney, Anne L. and Koornneef, Jan and Storchi-Bergmann, Thaisa},
  title = {The Dust Content and Opacity of Actively Star-forming Galaxies},
  journal = {The Astrophysical Journal},
  volume = {533},
  number = {2},
  pages = {682--695},
  year = {2000}
}

@inproceedings{liu2026denoising,
  author = {Liu, Shuhong and Ge, Xining and Gu, Ziying and Xu, Quanfeng and Gu, Lin and Cui, Ziteng and Chu, Xuangeng and Liu, Jun and Li, Dong and Harada, Tatsuya},
  title = {Denoising the Deep Sky: Physics-Based {CCD} Noise Formation for Astronomical Imaging},
  booktitle = {Computer Vision -- {ECCV} 2026},
  year = {2026},
  pages = {550--569}
}

@inproceedings{liu2026fluxflow,
  author = {Liu, Shuhong and Ge, Xining and Xu, Quanfeng and Cui, Ziteng and Li, Liuzhuozheng and Chang, Gengjia and Liu, Jun and Gu, Ziying and Li, Dong and Chu, Xuangeng and Gu, Lin and Harada, Tatsuya},
  title = {{FluxFlow}: Conservative Flow-Matching for Astronomical Image Super-Resolution},
  booktitle = {Advances in Neural Information Processing Systems (NeurIPS)},
  year = {2026}
}

@inproceedings{mildenhall2020nerf,
  author = {Mildenhall, Ben and Srinivasan, Pratul P. and Tancik, Matthew and Barron, Jonathan T. and Ramamoorthi, Ravi and Ng, Ren},
  title = {{NeRF}: Representing Scenes as Neural Radiance Fields for View Synthesis},
  booktitle = {Computer Vision -- {ECCV} 2020},
  year = {2020},
  pages = {405--421}
}

@inproceedings{Pumarola_2021_CVPR,
  author = {Pumarola, Albert and Corona, Enric and Pons-Moll, Gerard and Moreno-Noguer, Francesc},
  title = {{D-NeRF}: Neural Radiance Fields for Dynamic Scenes},
  booktitle = {Proceedings of the IEEE/CVF Conference on Computer Vision and Pattern Recognition},
  pages = {10318--10327},
  year = {2021}
}

@article{kerbl2023gaussians,
  author = {Kerbl, Bernhard and Kopanas, Georgios and Leimk{\"u}hler, Thomas and Drettakis, George},
  title = {{3D} Gaussian Splatting for Real-Time Radiance Field Rendering},
  journal = {ACM Transactions on Graphics},
  volume = {42},
  number = {4},
  year = {2023},
  pages = {1--14}
}

@inproceedings{luiten2024dynamic,
  author = {Luiten, Jonathon and Kopanas, Georgios and Leibe, Bastian and Ramanan, Deva},
  title = {Dynamic {3D} Gaussians: Tracking by Persistent Dynamic View Synthesis},
  booktitle = {2024 International Conference on {3D} Vision ({3DV})},
  year = {2024},
  pages = {800--809}
}

@inproceedings{Yang_2024_CVPR,
  author = {Yang, Ziyi and Gao, Xinyu and Zhou, Wen and Jiao, Shaohui and Zhang, Yuqing and Jin, Xiaogang},
  title = {Deformable {3D} Gaussians for High-Fidelity Monocular Dynamic Scene Reconstruction},
  booktitle = {Proceedings of the IEEE/CVF Conference on Computer Vision and Pattern Recognition},
  pages = {20331--20341},
  year = {2024}
}

@inproceedings{Lei_2025_CVPR,
  author = {Lei, Jiahui and Weng, Yijia and Harley, Adam W. and Guibas, Leonidas and Daniilidis, Kostas},
  title = {{MoSca}: Dynamic Gaussian Fusion from Casual Videos via {4D} Motion Scaffolds},
  booktitle = {Proceedings of the IEEE/CVF Conference on Computer Vision and Pattern Recognition},
  pages = {6165--6177},
  year = {2025}
}

@inproceedings{Wang_2025_ICCV,
  author = {Wang, Qianqian and Ye, Vickie and Gao, Hang and Zeng, Weijia and Austin, Jake and Li, Zhengqi and Kanazawa, Angjoo},
  title = {Shape of Motion: {4D} Reconstruction from a Single Video},
  booktitle = {Proceedings of the IEEE/CVF International Conference on Computer Vision},
  pages = {9660--9672},
  year = {2025}
}

@inproceedings{Zhou_2026_CVPR,
  author = {Zhou, Haoran and Lee, Gim Hee},
  title = {{MotionScale}: Reconstructing Appearance, Geometry, and Motion of Dynamic Scenes with Scalable {4D} Gaussian Splatting},
  booktitle = {Proceedings of the IEEE/CVF Conference on Computer Vision and Pattern Recognition},
  pages = {11860--11870},
  year = {2026}
}

@article{raissi2019pinn,
  author = {Raissi, Maziar and Perdikaris, Paris and Karniadakis, George E.},
  title = {Physics-informed neural networks: A deep learning framework for solving forward and inverse problems involving nonlinear partial differential equations},
  journal = {Journal of Computational Physics},
  volume = {378},
  pages = {686--707},
  year = {2019}
}

@inproceedings{hu2020difftaichi,
  author = {Hu, Yuanming and Anderson, Luke and Li, Tzu-Mao and Sun, Qi and Carr, Nathan and Ragan-Kelley, Jonathan and Durand, Fr{\'e}do},
  title = {{DiffTaichi}: Differentiable Programming for Physical Simulation},
  booktitle = {International Conference on Learning Representations},
  year = {2020}
}

@inproceedings{li2023pacnerf,
  author = {Li, Xuan and Qiao, Yi-Ling and Chen, Peter Yichen and Jatavallabhula, Krishna Murthy and Lin, Ming and Jiang, Chenfanfu and Gan, Chuang},
  title = {{PAC-NeRF}: Physics Augmented Continuum Neural Radiance Fields for Geometry-Agnostic System Identification},
  booktitle = {International Conference on Learning Representations},
  year = {2023}
}

@inproceedings{zhong2024springgaus,
  author = {Zhong, Licheng and Yu, Hong-Xing and Wu, Jiajun and Li, Yunzhu},
  title = {Reconstruction and Simulation of Elastic Objects with Spring-Mass {3D} Gaussians},
  booktitle = {Computer Vision -- {ECCV} 2024},
  year = {2024},
  pages = {407--423}
}

@inproceedings{jiang2025phystwin,
  author = {Jiang, Hanxiao and Hsu, Hao-Yu and Zhang, Kaifeng and Yu, Hsin-Ni and Wang, Shenlong and Li, Yunzhu},
  title = {{PhysTwin}: Physics-Informed Reconstruction and Simulation of Deformable Objects from Videos},
  booktitle = {Proceedings of the IEEE/CVF International Conference on Computer Vision},
  year = {2025},
  pages = {7219--7230}
}

@inproceedings{xie2024physgaussian,
  author = {Xie, Tianyi and Zong, Zeshun and Qiu, Yuxing and Li, Xuan and Feng, Yutao and Yang, Yin and Jiang, Chenfanfu},
  title = {{PhysGaussian}: Physics-Integrated {3D} Gaussians for Generative Dynamics},
  booktitle = {Proceedings of the IEEE/CVF Conference on Computer Vision and Pattern Recognition},
  pages = {4389--4398},
  year = {2024}
}

@inproceedings{zhang2024physdreamer,
  author = {Zhang, Tianyuan and Yu, Hong-Xing and Wu, Rundi and Feng, Brandon Y. and Zheng, Changxi and Snavely, Noah and Wu, Jiajun and Freeman, William T.},
  title = {{PhysDreamer}: Physics-Based Interaction with {3D} Objects via Video Generation},
  booktitle = {Computer Vision -- {ECCV} 2024},
  year = {2024},
  pages = {388--406}
}

@inproceedings{liu2025physflow,
  author = {Liu, Zhuoman and Ye, Weicai and Luximon, Yan and Wan, Pengfei and Zhang, Di},
  title = {Unleashing the Potential of Multi-modal Foundation Models and Video Diffusion for {4D} Dynamic Physical Scene Simulation},
  booktitle = {Proceedings of the IEEE/CVF Conference on Computer Vision and Pattern Recognition},
  year = {2025},
  pages = {11016--11025}
}

@inproceedings{zhao2025physsplat,
  author = {Zhao, Haoyu and Wang, Hao and Zhao, Xingyue and Fei, Hao and Wang, Hongqiu and Long, Chengjiang and Zou, Hua},
  title = {{PhysSplat}: Efficient Physics Simulation for {3D} Scenes via {MLLM}-Guided Gaussian Splatting},
  booktitle = {Proceedings of the IEEE/CVF International Conference on Computer Vision},
  year = {2025},
  pages = {5242--5252}
}

@inproceedings{lv2026physgm,
  author = {Lv, Chunji and Chen, Zequn and Di, Donglin and Zhang, Weinan and Li, Hao and Wei, Chen and Lei, Yinjie and Li, Changsheng},
  title = {{PhysGM}: Large Physical Gaussian Model for Feed-Forward {4D} Synthesis},
  booktitle = {Proceedings of the IEEE/CVF Conference on Computer Vision and Pattern Recognition},
  year = {2026},
  pages = {29855--29865}
}

@inproceedings{park2021nerfies,
  author = {Park, Keunhong and Sinha, Utkarsh and Barron, Jonathan T. and Bouaziz, Sofien and Goldman, Dan B and Seitz, Steven M. and Martin-Brualla, Ricardo},
  title = {Nerfies: Deformable Neural Radiance Fields},
  booktitle = {Proceedings of the IEEE/CVF International Conference on Computer Vision},
  year = {2021},
  pages = {5865--5874}
}

@article{park2021hypernerf,
  author = {Park, Keunhong and Sinha, Utkarsh and Hedman, Peter and Barron, Jonathan T. and Bouaziz, Sofien and Goldman, Dan B and Martin-Brualla, Ricardo and Seitz, Steven M.},
  title = {{HyperNeRF}: A Higher-Dimensional Representation for Topologically Varying Neural Radiance Fields},
  journal = {ACM Transactions on Graphics},
  volume = {40},
  number = {6},
  year = {2021},
  pages = {1--12}
}

@inproceedings{fridovich2023kplanes,
  author = {Fridovich-Keil, Sara and Meanti, Giacomo and Warburg, Frederik Rahb{\ae}k and Recht, Benjamin and Kanazawa, Angjoo},
  title = {{K-Planes}: Explicit Radiance Fields in Space, Time, and Appearance},
  booktitle = {Proceedings of the IEEE/CVF Conference on Computer Vision and Pattern Recognition},
  pages = {12479--12488},
  year = {2023}
}

@inproceedings{li2024spacetime,
  author = {Li, Zhan and Chen, Zhang and Li, Zhong and Xu, Yi},
  title = {Spacetime Gaussian Feature Splatting for Real-Time Dynamic View Synthesis},
  booktitle = {Proceedings of the IEEE/CVF Conference on Computer Vision and Pattern Recognition},
  pages = {8508--8520},
  year = {2024}
}

@inproceedings{huang2024scgs,
  author = {Huang, Yi-Hua and Sun, Yang-Tian and Yang, Ziyi and Lyu, Xiaoyang and Cao, Yan-Pei and Qi, Xiaojuan},
  title = {{SC-GS}: Sparse-Controlled Gaussian Splatting for Editable Dynamic Scenes},
  booktitle = {Proceedings of the IEEE/CVF Conference on Computer Vision and Pattern Recognition},
  pages = {4220--4230},
  year = {2024}
}

@inproceedings{zhang2026continuous,
  author = {Zhang, Xuankai and Xiao, Junjin and Huang, Shangwei and Zheng, Wei-shi and Zhang, Qing},
  title = {Learning Explicit Continuous Motion Representation for Dynamic Gaussian Splatting from Monocular Videos},
  booktitle = {Proceedings of the IEEE/CVF Conference on Computer Vision and Pattern Recognition},
  pages = {33291--33300},
  year = {2026}
}

@inproceedings{chen2018neuralode,
  author = {Chen, Ricky T. Q. and Rubanova, Yulia and Bettencourt, Jesse and Duvenaud, David K.},
  title = {Neural Ordinary Differential Equations},
  booktitle = {Advances in Neural Information Processing Systems},
  volume = {31},
  year = {2018}
}

@inproceedings{greydanus2019hamiltonian,
  author = {Greydanus, Samuel and Dzamba, Misko and Yosinski, Jason},
  title = {Hamiltonian Neural Networks},
  booktitle = {Advances in Neural Information Processing Systems},
  volume = {32},
  year = {2019}
}

@article{chu2022physics,
  author = {Chu, Mengyu and Liu, Lingjie and Zheng, Quan and Franz, Aleksandra and Seidel, Hans-Peter and Theobalt, Christian and Zayer, Rhaleb},
  title = {Physics Informed Neural Fields for Smoke Reconstruction with Sparse Data},
  journal = {ACM Transactions on Graphics},
  volume = {41},
  number = {4},
  year = {2022},
  pages = {1--14}
}

@inproceedings{qiao2022neuphysics,
  author = {Qiao, Yi-Ling and Gao, Alexander and Lin, Ming},
  title = {{NeuPhysics}: Editable Neural Geometry and Physics from Monocular Videos},
  booktitle = {Advances in Neural Information Processing Systems},
  volume = {35},
  year = {2022},
  pages = {12841--12854}
}

@inproceedings{du2025gaussfluids,
  author = {Du, Feilong and Zhang, Yalan and Ji, Yihang and Wang, Xiaokun and Yao, Chao and Kosinka, Jiri and Frey, Steffen and Telea, Alexandru and Ban, Xiaojuan},
  title = {{GaussFluids}: Reconstructing Lagrangian Fluid Particles from Videos via Gaussian Splatting},
  booktitle = {Pacific Graphics Conference Papers, Posters, and Demos},
  year = {2025}
}

@inproceedings{li2026ngff,
  author = {Li, Shiqian and Shen, Ruihong and Ni, Junfeng and Pan, Chang and Zhang, Chi and Zhu, Yixin},
  title = {Learning Physics-Grounded {4D} Dynamics with Neural Gaussian Force Fields},
  booktitle = {International Conference on Learning Representations},
  year = {2026}
}

@article{blandhawthorn2016galaxy,
  title   = {The Galaxy in Context: Structural, Kinematic, and Integrated Properties},
  author  = {Bland-Hawthorn, Joss and Gerhard, Ortwin},
  journal = {Annual Review of Astronomy and Astrophysics},
  volume  = {54}, pages = {529--596}, year = {2016},
  doi     = {10.1146/annurev-astro-081915-023441}
}

@article{li2021m31,
  title   = {A Sub-2\% Distance to {M31} from Photometrically Homogeneous Near-Infrared
             {Cepheid} Period--Luminosity Relations Measured with the {Hubble Space Telescope}},
  author  = {Li, Siyang and Riess, Adam G. and Busch, Michael P. and Casertano, Stefano
             and Macri, Lucas M. and Yuan, Wenlong},
  journal = {The Astrophysical Journal},
  volume  = {920}, number = {2}, pages = {84}, year = {2021},
  doi     = {10.3847/1538-4357/ac1597}
}

@article{chemin2009m31,
  title   = {{H\,I} Kinematics and Dynamics of {Messier 31}},
  author  = {Chemin, Laurent and Carignan, Claude and Foster, Tyler},
  journal = {The Astrophysical Journal},
  volume  = {705}, number = {2}, pages = {1395--1415}, year = {2009},
  doi     = {10.1088/0004-637X/705/2/1395}
}

@article{brunthaler2005m33,
  title   = {The Geometric Distance and Proper Motion of the {Triangulum Galaxy (M33)}},
  author  = {Brunthaler, Andreas and Reid, Mark J. and Falcke, Heino
             and Greenhill, Lincoln J. and Henkel, Christian},
  journal = {Science},
  volume  = {307}, number = {5714}, pages = {1440--1443}, year = {2005},
  doi     = {10.1126/science.1108342}
}

@article{sohn2012m31,
  title   = {The {M31} Velocity Vector. {I}. {Hubble Space Telescope} Proper-Motion Measurements},
  author  = {Sohn, Sangmo Tony and Anderson, Jay and van der Marel, Roeland P.},
  journal = {The Astrophysical Journal},
  volume  = {753}, number = {1}, pages = {7}, year = {2012},
  doi     = {10.1088/0004-637X/753/1/7}
}

@String(CVPR= {IEEE Conf. Comput. Vis. Pattern Recog.})

@String(ECCV= {Eur. Conf. Comput. Vis.})

@String(CVPR  = {CVPR})

@String(ECCV  = {ECCV})
}


\end{document}